\documentclass{article}

\usepackage{microtype}
\usepackage{graphicx}
\usepackage{subfigure}
\usepackage{booktabs} 

\usepackage[utf8]{inputenc} 
\usepackage[T1]{fontenc}    
\usepackage{hyperref}       
\usepackage{url}            
\usepackage{amsfonts}       
\usepackage{nicefrac}       
\usepackage{xcolor}         

\usepackage{makecell} 
\usepackage{amsmath}
\usepackage{amssymb}
\usepackage{mathtools}
\usepackage{amsthm}

\usepackage{url}
\usepackage{bbm}
\usepackage{amssymb}
\usepackage{enumitem}
\usepackage{algorithm, algorithmic}
\usepackage{wrapfig}
\usepackage{mathrsfs}
\usepackage{multirow}
\usepackage{amsthm}
\usepackage{caption}
\usepackage{enumitem}
\usepackage{pifont}

\usepackage[accepted]{icml2021}

\icmltitlerunning{Confident or Seek Stronger: Exploring Uncertainty-Based On-device LLM Routing From Benchmarking to Generalization}

\begin{document}

\twocolumn[
\icmltitle{Confident or Seek Stronger: Exploring Uncertainty-Based \\ On-device LLM  Routing From Benchmarking to Generalization}  



\icmlsetsymbol{equal}{*}

\begin{icmlauthorlist}
\icmlauthor{Yu-Neng Chuang}{equal,to}
\icmlauthor{Leisheng Yu}{equal,to}
\icmlauthor{Guanchu Wang}{to}
\icmlauthor{Lizhe Zhang}{sd}
\icmlauthor{Zirui Liu}{too} \\
\icmlauthor{Xuanting Cai}{ed}
\icmlauthor{Yang Sui}{to}
\icmlauthor{Vladimir Braverman}{goo}
\icmlauthor{Xia Hu}{to}
\end{icmlauthorlist}

\icmlaffiliation{to}{Rice University}
\icmlaffiliation{goo}{Johns Hopkins University}
\icmlaffiliation{too}{University of Minnesota}
\icmlaffiliation{ed}{Meta AI}
\icmlaffiliation{sd}{University of California, San Diego }

\icmlcorrespondingauthor{Yu-Neng Chuang}{ynchuang@rice.edu}
\icmlcorrespondingauthor{Leisheng Yu}{ly50@rice.edu}
\icmlcorrespondingauthor{Xia Hu}{xia.hu@rice.edu}

\icmlkeywords{Machine Learning, ICML}
\vskip 0.3in
]



\printAffiliationsAndNotice{\icmlEqualContribution}
\vspace{-0.5cm}

\begin{abstract}
Large language models (LLMs) are increasingly deployed and democratized on edge devices. To improve the efficiency of on-device deployment, small language models (SLMs) are often adopted due to their efficient decoding latency and reduced energy consumption.
However, these SLMs often generate inaccurate responses when handling complex queries. One promising solution is uncertainty-based SLM routing, offloading high-stakes queries to stronger LLMs when resulting in low-confidence responses on SLM. This follows the principle of \textit{If you lack confidence, seek stronger support} to enhance reliability. Relying on more powerful LLMs is yet effective but increases invocation costs. Therefore, striking a routing balance between efficiency and efficacy remains a critical challenge. Additionally, efficiently generalizing the routing strategy to new datasets remains under-explored. In this paper, we conduct a comprehensive investigation into \textit{\textbf{benchmarking and generalization of uncertainty-driven routing strategies from SLMs to LLMs over 1500+ settings}}. 
Our findings highlight: \textit{First}, uncertainty-correctness alignment in different uncertainty quantification (UQ) methods significantly impacts routing performance. \textit{Second}, uncertainty distributions depend more on both the specific SLM and the chosen UQ method, rather than downstream data.
Building on the insight, we propose a calibration data conp sstruction pipeline and open-source a constructed hold-out set to enhance routing generalization on new downstream scenarios. The experimental results indicate calibration data effectively bootstraps routing performance without any new data. 


\end{abstract}

\section{Introduction}

Large language models (LLMs) deployment on edge devices have gained increasing attention in recent years, primarily due to their potential for low-latency, privacy-preserving inference. Given the computational and memory constraints of edge devices, small language models (SLMs) (e.g., Phi2-mini~\cite{phi2} or Llama3.2-3B~\cite{touvron2023llama} are designed for resource-efficient deployment, particularly on devices such as smartphones and wearable devices. Their overarching goal is to democratize the deployment of LMs, making it accessible and affordable to users across diverse settings and at any time~\cite{lu2024small, zhou2023mini, zhao2023survey}.
However, these SLMs often lack the robustness and scalability of LLMs~\cite{chen2024role} (e.g., GPT-4o~\cite{achiam2023gpt} and Llama-3.1-405B), especially when faced with diverse and complex input queries under the deployment on edge devices, which eventually degrade the overall performance.
This limitation raises a critical need for exploring the solutions to increase the response reliability of SLMs.

To mitigate this unreliability, a line of work proposes to partially offload challenging and complex queries from SLMs to LLMs~\cite{chuang2024learning, ong2024routellm, hu2024routerbench, stripelis2024polyrouter}. A hybrid system is then established to wisely route the queries from SLMs and seek more reliable and deterministic responses from stronger LLMs.
Although LLMs can exhibit superior performance, they incur high maintenance and inference costs given the large scale of model size and their infrastructure (i.e., a single NVIDIA A$100$ GPU can cost approximately \$2,000 per month for deployment). 
Inaccurate routing by SLMs increases the volume of queries forwarded to LLMs, necessitating greater bandwidth allocation for maintaining the service of LLMs. As a result, operational costs and budgetary requirements rise accordingly, especially when continuous deployment is required.
Hence, developing an effective routing strategy is crucial for fully deploying SLMs~\cite{ong2024routellm,stripelis2024polyrouter,chuang2024learning}, as it both enhances response reliability and reduces the costs associated with services and data transmission.


Leveraging SLMs’ self-uncertainty estimation emerges as a robust strategy for enhancing routing effectiveness~\cite{chuang2024learning, dinghybrid}. By relying on the self-assessed uncertainty, the system can better decide whether to handle a query locally or delegate it to a larger model without the aid of extra routers, ensuring that only queries deemed unreliable by the SLMs are routed to LLMs. As a result, the uncertainty-based routing approach not only generalizes well to new datasets, as only self-assessed information from SLM is needed, but it also reduces the high operational costs associated with accurately running LLMs. To this end, we aim to explore two open and nontrivial research questions for uncertainty-based SLM routing:

\textbf{\textit{1) What is the best practice of uncertainty estimation for query routing from SLMs to LLMs?}} 
In this research question, we benchmark the uncertainty-correctness alignment of each uncertainty quantification (UQ) method under their impact to SLM routing. A good alignment is a key factor for successful routing decisions, as any misalignment can cause unnecessary offloading with extra cost. However, SLMs may struggle to provide reliable uncertainty estimates~\cite{huang2023look, detommaso2024multicalibration, wightman2023strength}, making them less effective as indicators for query routing. Thus, we benchmark the alignment between uncertainty and correctness, paving the insights for establishing more effective routing strategies\footnote{For the convenience, we interchangeably use uncertainty and confidence, where low uncertainty refers to high confidence.}.

\textbf{\textit{2) What is the best practice to initially establish an effective routing strategy when generalizing to new datasets?}}
In this research question, we explore how to generalize routing strategies to new datasets. Existing approaches~\cite{ong2024routellm,hu2024routerbench} rely on sufficient new downstream data to make routing decisions for optimal performance-cost trade-offs, but this process is time-consuming and labor-intensive. Broadly speaking, collecting and analyzing full downstream datasets under varying SLM configurations can be prohibitively costly, delaying implementation, which is not practical in real-world scenarios. 
This delay is particularly problematic in high-stakes scenarios, such as medical wearable devices, where reliability is critical, and inaccuracies are unacceptable even in early deployment stages. Based on our findings, we provide a data construction pipeline to enhance the generalization of establishing routing in new downstream scenarios. A generated calibration dataset as a data-agnostic hold-out set enables the estimation of effective routing decisions without requiring extensive new data. We further benchmark the benefits of this calibration dataset, demonstrating the one's generalization ability to new datasets.

\begin{table*}[t!]
  \centering
  \caption{Uncertainty quantification (UQ) methods evaluated in our benchmark. “Model Access” specifies whether a method views the LM’s weights/logits (white-box) or only its generated output (black-box). “Require Training?” indicates if additional training is needed. See Subsection~\ref{sec:uqs} for taxonomy details and Subsection~\ref{sec:benchset} for method descriptions.}
  \vspace{0.25cm}
  \label{tab:uq_overveiw}
  \resizebox{.95\textwidth}{!}{%
  \begin{tabular}{lcccc} 
    \toprule
    Uncertainty Quantification (UQ) Methods & Taxonomy & Model Access & Require Training? \\
    \midrule
    Average Token Prob~\cite{mahaut2024factual} & Token/sequence probabilities & White-box & No \\
    $p(\text{True})$~\cite{kadavath2022language} & Token/sequence probabilities & White-box & No \\
    Perplexity~\cite{fomicheva2020unsupervised} & Token/sequence probabilities & White-box & No \\
    Jaccard Degree~\cite{lin2023generating} & Output consistency & Black-box & No \\
    Verbalization-1s~\cite{xiongcan, tian2023just} & Verbalized uncertainty & Black-box & No \\
    Verbalization-2s~\cite{tian2023just} & Verbalized uncertainty & Black-box & No \\
    \midrule
    Trained Probe~\cite{azaria2023internal, kadavath2022language, mahaut2024factual} & Uncertainty probe & White-box & Yes\\
    OOD Probe~\cite{kadavath2022language, mahaut2024factual} & Uncertainty probe & White-box & Yes\\
    \bottomrule
  \end{tabular}}
\end{table*}

This work offers an accessible and reproducible pipeline for uncertainty-based routing from benchmarking to generalization. Our main contributions are summarized as follows:
\begin{itemize}[leftmargin=*]
    \item \textbf{Comprehensive benchmarking and detailed analysis:} This benchmark evaluates $8$ UQ methods across $14$ datasets to examine the alignment between uncertainty and correctness in routing tasks. We incorporate $8$ SLMs and $2$ LLMs to emulate real-world deployment scenarios. We then delve into key observations from the extensive results and conclude the insights for developing uncertainty-based SLM routing.

    \item \textbf{Calibration data for generalization to new data:} Building on our benchmarking pipeline, we introduce a calibration data construction pipeline designed to improve routing generalization in new downstream scenarios. Empirical results show that this calibration data generalizes effectively to new datasets \textit{without relying on any new downstream data}.
    
\end{itemize}

\section{Reviewing Different Schools of Uncertainty Quantification and LLM Routing}
\subsection{Uncertainty Quantification for LMs} \label{sec:uqs}
Uncertainty quantification methods estimate a model’s confidence in its predictions~\cite{houdecomposing}. For traditional classification and regression, uncertainty estimation is well-established~\cite{gal2016uncertainty}. However, for LLMs generating free-form responses to complex queries, estimating uncertainty is more challenging because the output space can grow exponentially with vocabulary size, and each sequence spans multiple tokens ~\cite{fadeeva2023lm}. Existing uncertainty quantification approaches for LLMs can be grouped into the following four categories.

\noindent\textbf{Via Verbalizing Uncertainty.} This line of work prompts language models to report linguistic confidence~\cite{mahaut2024factual, mielke2022reducing}. To enable LMs to verbalize confidence, researchers have proposed fine-tuning them to express uncertainty~\cite{lin2022teaching} or teaching them to verbalize confidence through in-context learning~\cite{dong2024survey}. Verbalized confidence can take the form of linguistic expressions of uncertainty or numerical scores~\cite{geng2024survey}. Multiple studies find that LLMs tend to be overconfident when reporting confidence~\cite{xiongcan, tian2023just}. To mitigate this overconfidence, prompting strategies such as multi-step elicitation, top-$k$, and Chain-of-Thought~\cite{wei2022chain} have been explored~\cite{tian2023just}. Sampling multiple response-confidence pairs and designing more effective aggregation strategies can also help mitigate overconfidence~\cite{xiongcan}. Moreover, \citeauthor{tian2023just} report that verbalized confidence is typically better calibrated than the model’s conditional probabilities.

\noindent\textbf{Via Analyzing Token/Sequence Probabilities.}
This line of research derives confidence scores from model logits for output tokens~\cite{geng2024survey, huang2023look, jiang2021can}. The confidence of a generated sequence is computed by aggregating the log-probabilities of its tokens. Common aggregation strategies include arithmetic average, minimum, perplexity, and average entropy~\cite{fadeeva2023lm, fomicheva2020unsupervised, vazhentsev2023efficient}. Because not all tokens in a sequence equally reflect semantic content, SAR reweights token likelihoods to emphasize more meaningful tokens~\cite{duan2024shifting}. However, different surface realizations of the same claim can yield different probabilities, implying that the calculated confidence reflects how a claim is articulated rather than the claim itself~\cite{mahaut2024factual}. To combine LM self-assessment with token probabilities, $p(\text{True})$ is proposed: the model is asked whether its generated response is correct, and the probabilities of True/False tokens serve as the confidence score~\cite{kadavath2022language, tian2023just}.

\noindent\textbf{Via Gauging Output Consistency.} This line of research (e.g., SelfCheckGPT~\cite{manakul2023selfcheckgpt}) assumes that high-confidence LLMs produce consistent outputs~\cite{mahaut2024factual}. A typical approach samples $m$ responses for a given input query, measures inter-response similarity, and calculates a confidence score from meaning diversity~\cite{fadeeva2023lm}. Common ways to measure pairwise similarity include Natural Language Inference (NLI) and Jaccard similarity~\cite{geng2024survey}. Consistency is then assessed by analyzing the similarity matrix, for instance, by counting semantic sets, summing eigenvalues of the graph Laplacian, examining the degree matrix, or computing eccentricity~\cite{lin2023generating}. Because different sentences can express the same meaning, semantic entropy~\cite{kuhn2023semantic} first clusters responses by semantic equivalence before measuring consistency.

\noindent\textbf{Via Training Uncertainty Probes.} This approach trains classifiers to predict whether an LLM will arrive at the correct answer for a particular query, using predicted probabilities as confidence scores~\cite{geng2024survey}. Training data is often obtained by sampling multiple answers per question at a fixed temperature and labeling each for correctness~\cite{kadavath2022language}. A probe (commonly a multi-layer perceptron) then takes hidden states as inputs to predict correctness~\cite{azaria2023internal, li2024inference}. Because in-domain training data is not always available, Contrast-Consistent Search trains probes unsupervisedly by maximizing representation distances between contradictory answers on Yes/No questions~\cite{burns2022discovering}. Furthermore, whether probes trained on out-of-distribution data remain effective is still under debate~\cite{kadavath2022language, mahaut2024factual, kuhn2023semantic, liu2024uncertainty}.


\begin{figure*}
\centering
    \includegraphics[width=0.95\textwidth]{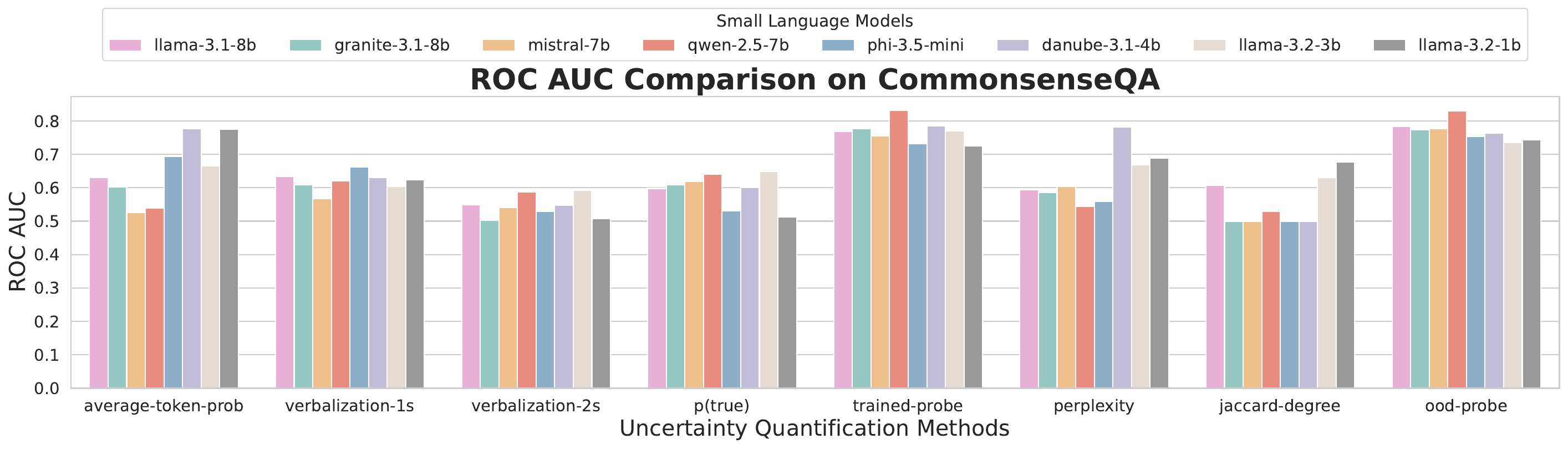} 
    \vspace{-3mm}
    \caption{The ROC AUC scores measure the alignment between confidence and correctness across different SLMs and uncertainty quantification methods on CommonsenseQA. A higher ROC AUC indicates a stronger alignment.}
\label{fig:uq_align} 
\end{figure*}

\subsection{LLM Routing} 
In query-routing scenarios, recent approaches train additional classifiers to direct queries to different SLMs or LLMs based on historical performance metrics and user feedback data~\cite{dinghybrid, ong2024routellm, stripelis2024polyrouter, jiang2024mixtral, zhao2024eagle}. For instance, RouterBench~\cite{hu2024routerbench} collects inference outputs from selected LLMs to aid in the development of routing classifiers. However, these methods face significant challenges when encountering new downstream tasks, as such data falls outside the distribution of the existing training data. This limitation makes them less practical for real-world scenarios, such as on personal edge device deployment, where adaptability to unseen conditions is crucial.
Our work focuses on how to establish routing systems between SLMs and LLMs and generalize to new downstream tasks. In this manner, uncertainty-based routing is an appropriate solution to overcome these challenges, as uncertainty is directly extracted from SLMs themselves. Furthermore, we propose a calibration data construction pipeline to initialize a routing system that generalizes to unseen datasets.

\section{Benchmarking Uncertainty-based SLM Routing}
In this section, we systematically evaluate $8$ SLMs and $2$ LLMs on $14$ datasets using $8$ UQ methods (see Table~\ref{tab:uq_overveiw}) for uncertainty-based SLM routing. This section details the datasets, models, and UQ methods, followed by several key findings and practical considerations. All experiments are conducted on four $80$GB NVIDIA A$100$ GPUs.

\subsection{Benchmark Coverage and Setup} \label{sec:benchset}
\noindent\textbf{Language Models.} We select $8$ open-source SLMs spanning $1$--$8$B parameters from multiple affiliations: Llama-3.2-1B-Instruct~\cite{meta2024llama}, Llama-3.2-3B-Instruct~\cite{meta2024llama}, Phi-3.5-mini-instruct~\cite{abdin2024phi}, danube3.1-4b-chat~\cite{pfeiffer2024h2o}, Mistral-7B-Instruct-v0.3~\cite{jiang2023mistral}, Qwen2.5-7B-Instruct~\cite{qwen2}, Llama-3.1-8B-Instruct~\cite{dubey2024llama}, and granite-3.1-8b-instruct~\cite{granite2024granite}. Three are from Meta, and the rest are contributed by Microsoft, H2O.ai, Mistral AI, Alibaba, and IBM. All adopt a decoder-only Transformer architecture and are available through Hugging Face. We include one open-source LLM (Llama-3.1-70B-Instruct~\cite{dubey2024llama}) and one proprietary API-based counterpart (GPT-4o mini~\cite{hurst2024gpt}).

\noindent\textbf{Datasets.} Experiments span $14$ datasets from four domains: \emph{(1) Mathematical Reasoning} (AQuA~\cite{ling2017program}, GSM8K~\cite{cobbe2021training}, MultiArith~\cite{roy2016solving}, SVAMP~\cite{patel2021nlp}), \emph{(2) Commonsense Reasoning} (CommonsenseQA~\cite{talmor-etal-2019-commonsenseqa}, HellaSwag~\cite{zellers2019hellaswag}, OpenBookQA~\cite{mihaylov2018can}, PIQA~\cite{bisk2020piqa}, TruthfulQA~\cite{lin2021truthfulqa}, WinoGrande~\cite{sakaguchi2021winogrande}, BoolQ~\cite{clark2019boolq}, Social IQa~\cite{sap2019social}), \emph{(3) Conversational and Contextual Understanding} (CoQA~\cite{reddy-etal-2019-coqa}), and \emph{(4) Problem Solving} (MMLU~\cite{hendrycksmeasuring}). These cover free-form, multiple-choice, and True/False question answering and are available via Hugging Face. Table~\ref{tab:datasets} in Appendix~\ref{app:data} provides further details.

\noindent\textbf{UQ Methods and Hyperparameters.} We evaluate $8$ approaches from the four categories in Section~\ref{sec:uqs}. \emph{(1) Average token probability} uses the probability of the chosen option token (e.g., ``A'') for multiple-choice tasks or the mean probability of all generated tokens for free-form tasks. \emph{(2) Perplexity} is computed for a sequence of $N$ output tokens $\{y_i\}_{i=1}^N$ with probabilities $\{p(y_i)\}_{i=1}^N$ as $\mathrm{exp}\!\bigl(\tfrac{1}{N}\sum_{i=1}^{N}\ln p(y_i)\bigr)$, and its reciprocal serves as the confidence score. \emph{(3) $p(\text{True})$} is a method where the LM first outputs an answer, then evaluates the generated response using only “True” or “False.” The probabilities for these two tokens are normalized to sum to $1$, and the probability of “True” is used as confidence. \emph{(4) Verbalized confidence in a single response} (denoted as verbalization-1s) prompts the model to output both the answer and numeric confidence in one step. \emph{(5) Verbalized confidence in the second round} (denoted as verbalization-2s) obtains the confidence in a separate, follow-up query after the model has provided an answer. \emph{(6) The degree matrix} (denoted as jaccard-degree) generates $m=5$ samples (temperature $1.0$) for one query, computes pairwise Jaccard similarities, and sets confidence to $\mathrm{trace}(mI - D)/m^2$, where $D$ is the degree matrix. \emph{(7) Trained probe} is a four-layer MLP with LeakyReLU activations, trained on a fixed subsample of the in-domain training set for each dataset, taking as input the hidden states from the eighth-to-last transformer layer. We train for $20$ epochs (learning rate $5\times10^{-4}$). \emph{(8) Trained probe on out-of-distribution data} (denoted as ood-probe) is identical in architecture but trained on all other datasets. For instance, if AQuA is evaluated, the ood-probe is trained on the remaining $13$ datasets ($20$ epochs, learning rate $1\times10^{-4}$).

For verbalization-based methods, we discard queries when the model does not follow instructions to produce a confidence score. For free-form question answering, we use GPT-4o mini to evaluate whether a response is essentially equivalent to the ground truth answer~\cite{zheng2023judging}.

\begin{figure*}[!t]
    \centering
    \subfigure[Granite-3.1-8B]{
    \centering
    \begin{minipage}[t]{0.23\linewidth}
	    \includegraphics[width=1.05\linewidth]{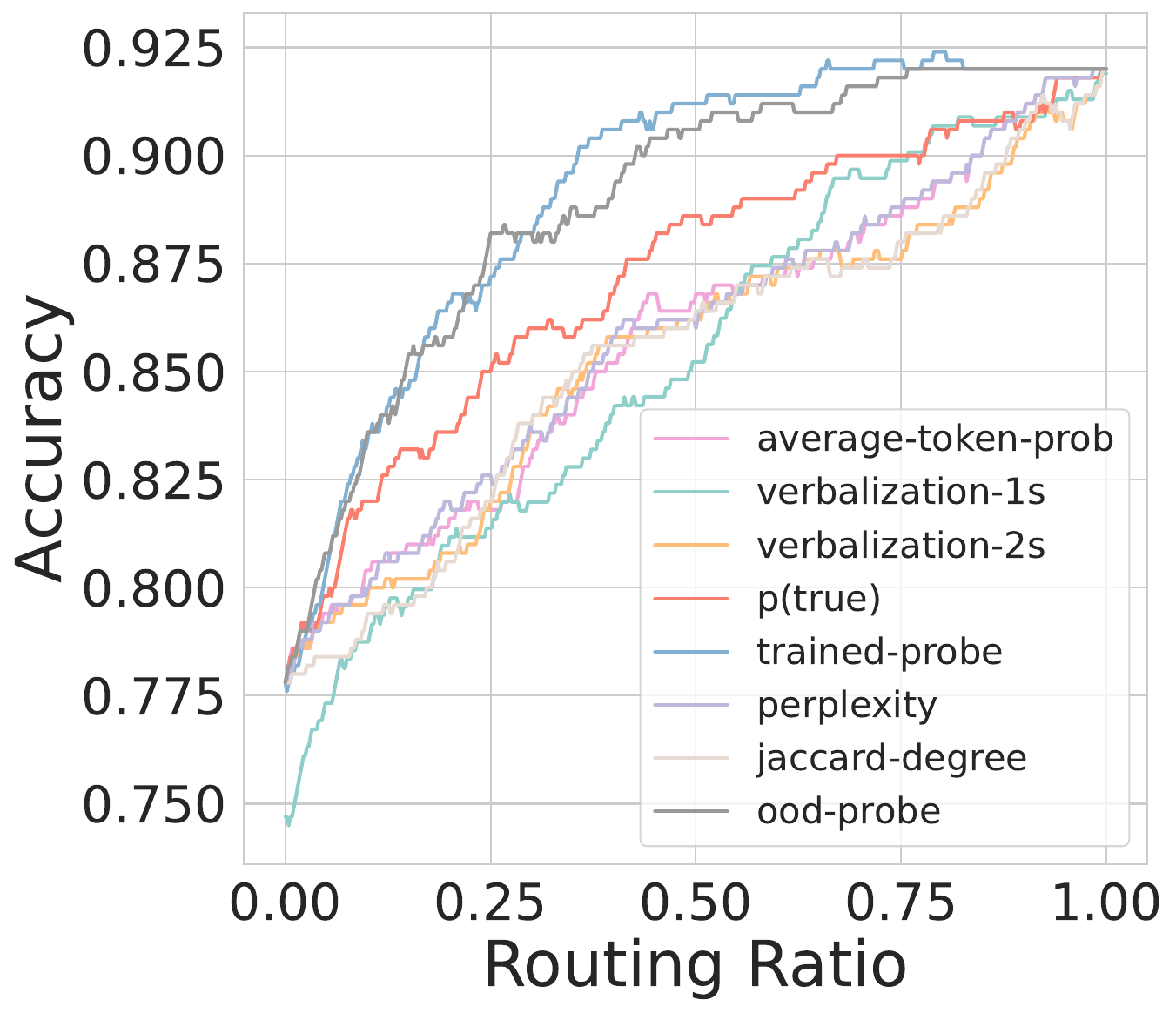}
    \end{minipage}%
    }
    \subfigure[Llama-3.2-3B]{
    \centering
    \begin{minipage}[t]{0.23\linewidth}
	    \includegraphics[width=1.05\linewidth]{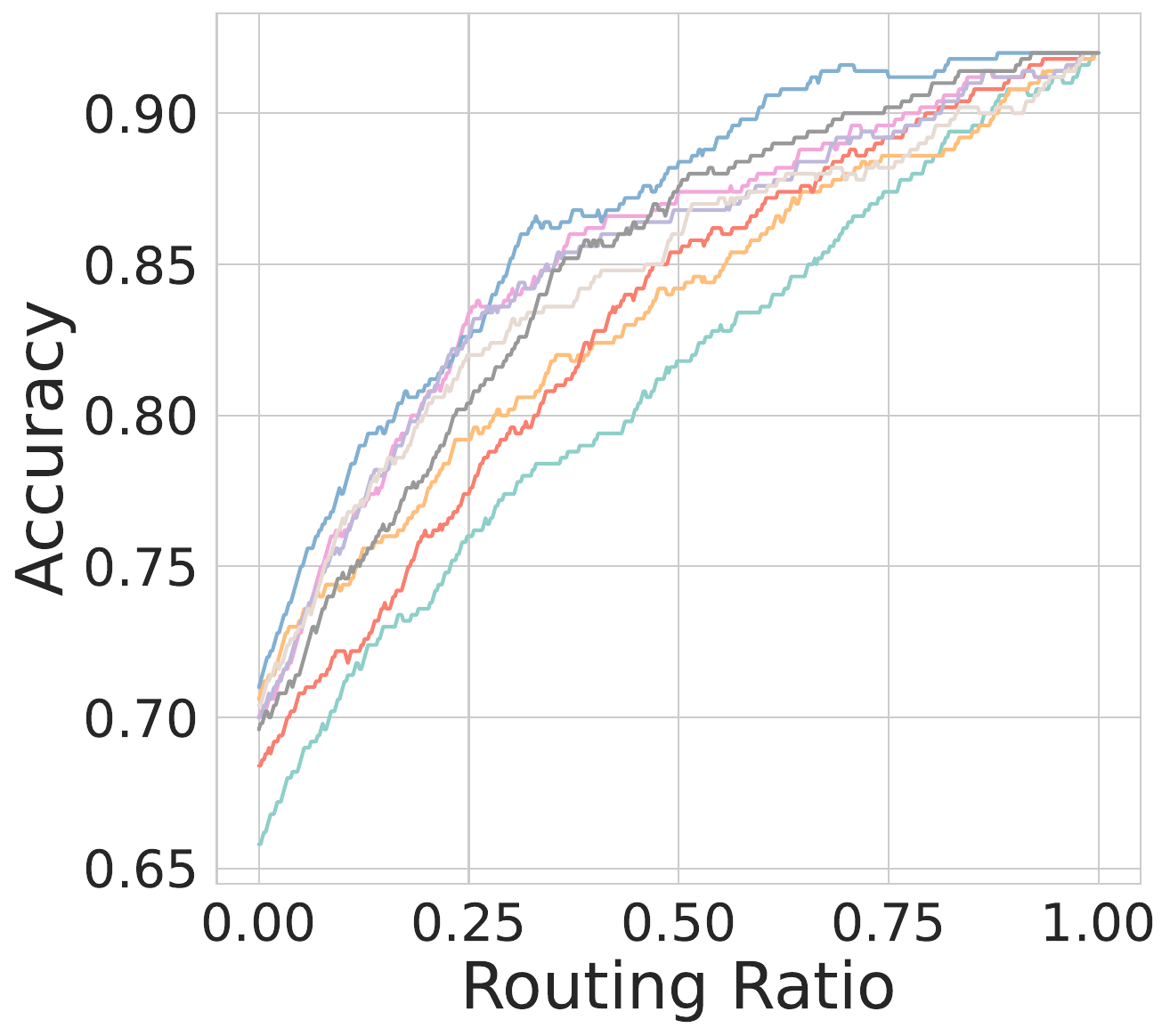}
    \end{minipage}%
    }
    \subfigure[Phi-3.5-mini]{
    \centering
    \begin{minipage}[t]{0.23\linewidth}
	    \includegraphics[width=1.05\linewidth]{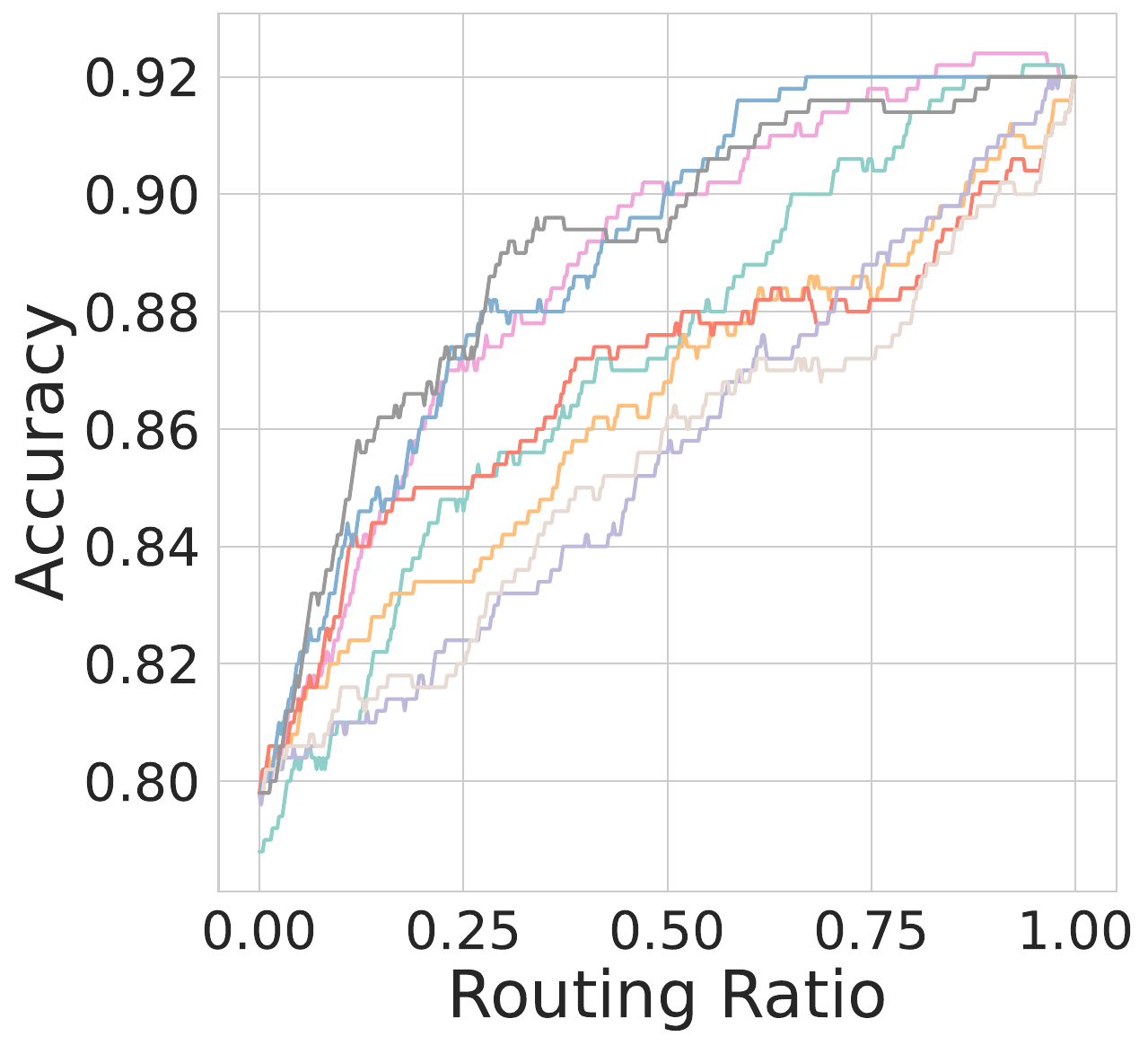}
    \end{minipage}%
    }
    \subfigure[Qwen-2.5-7B]{
    \centering
    \begin{minipage}[t]{0.23\linewidth}
	    \includegraphics[width=1.05\linewidth]{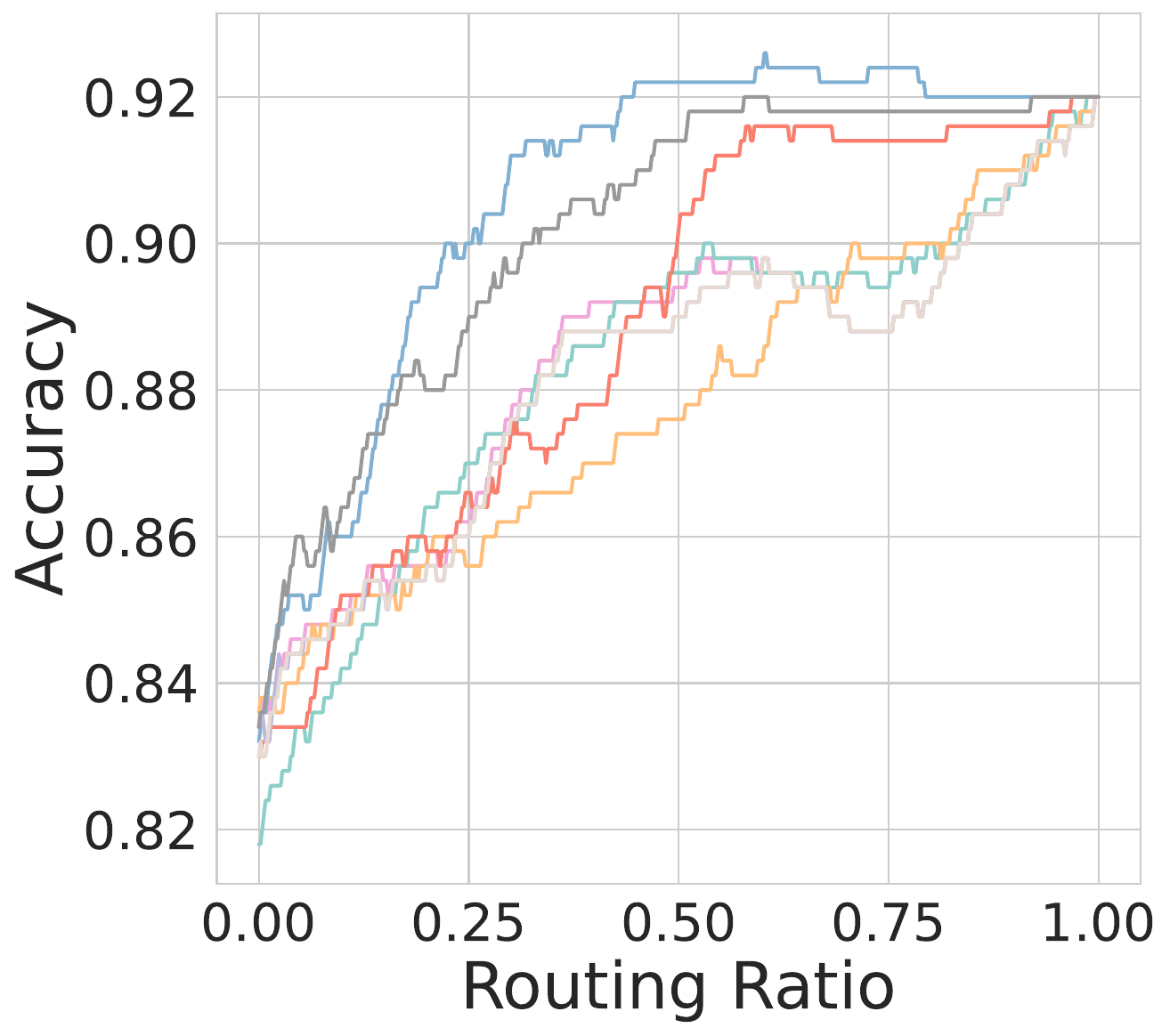}
    \end{minipage}%
    }
    \vspace{-0.1cm}
    \subfigure[Granite-3.1-8B]{
    \centering
    \begin{minipage}[t]{0.23\linewidth}
	    \includegraphics[width=1.05\linewidth]{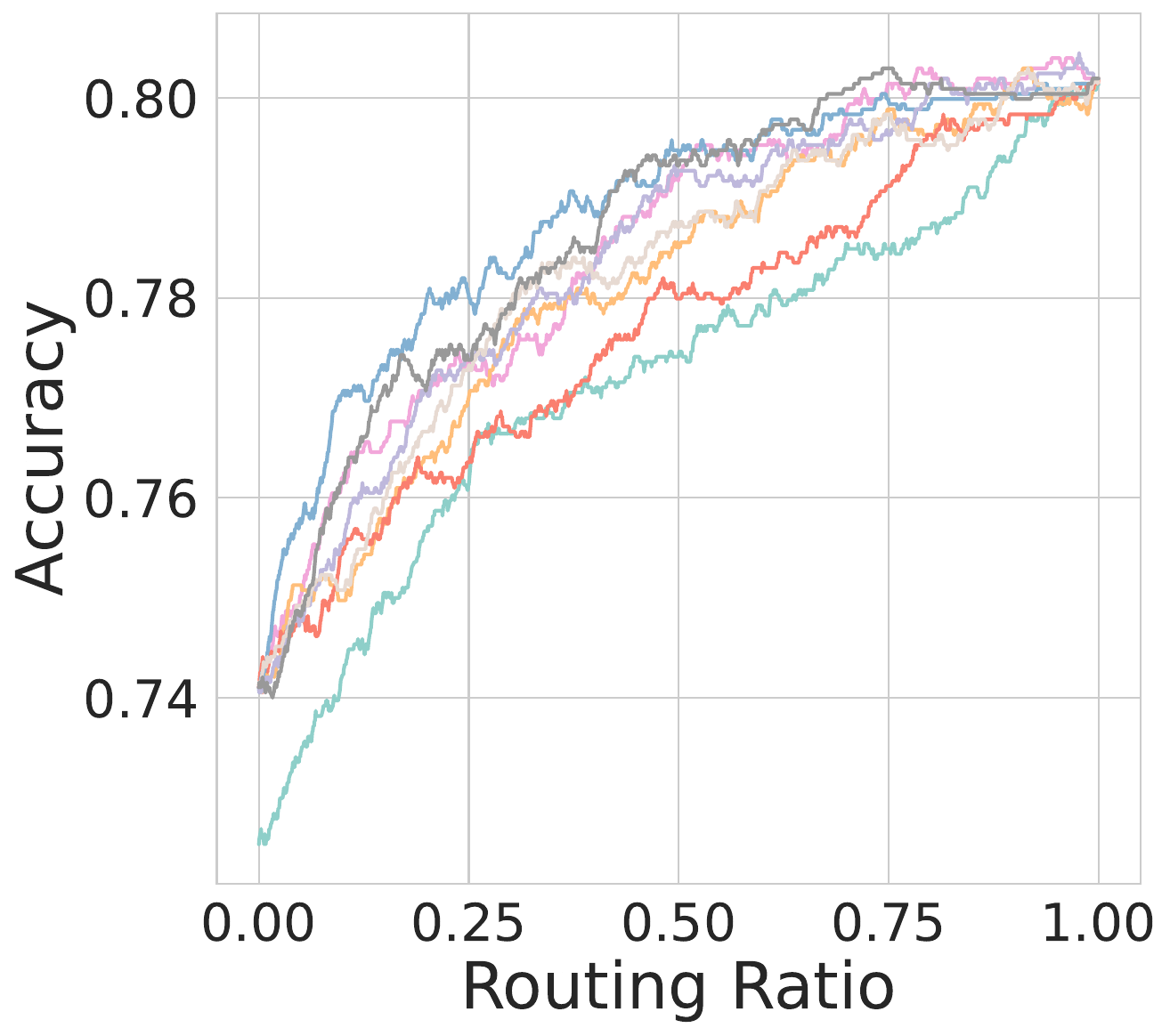}
    \end{minipage}%
    }
    \subfigure[Llama-3.2-3B]{
    \centering
    \begin{minipage}[t]{0.23\linewidth}
	    \includegraphics[width=1.05\linewidth]{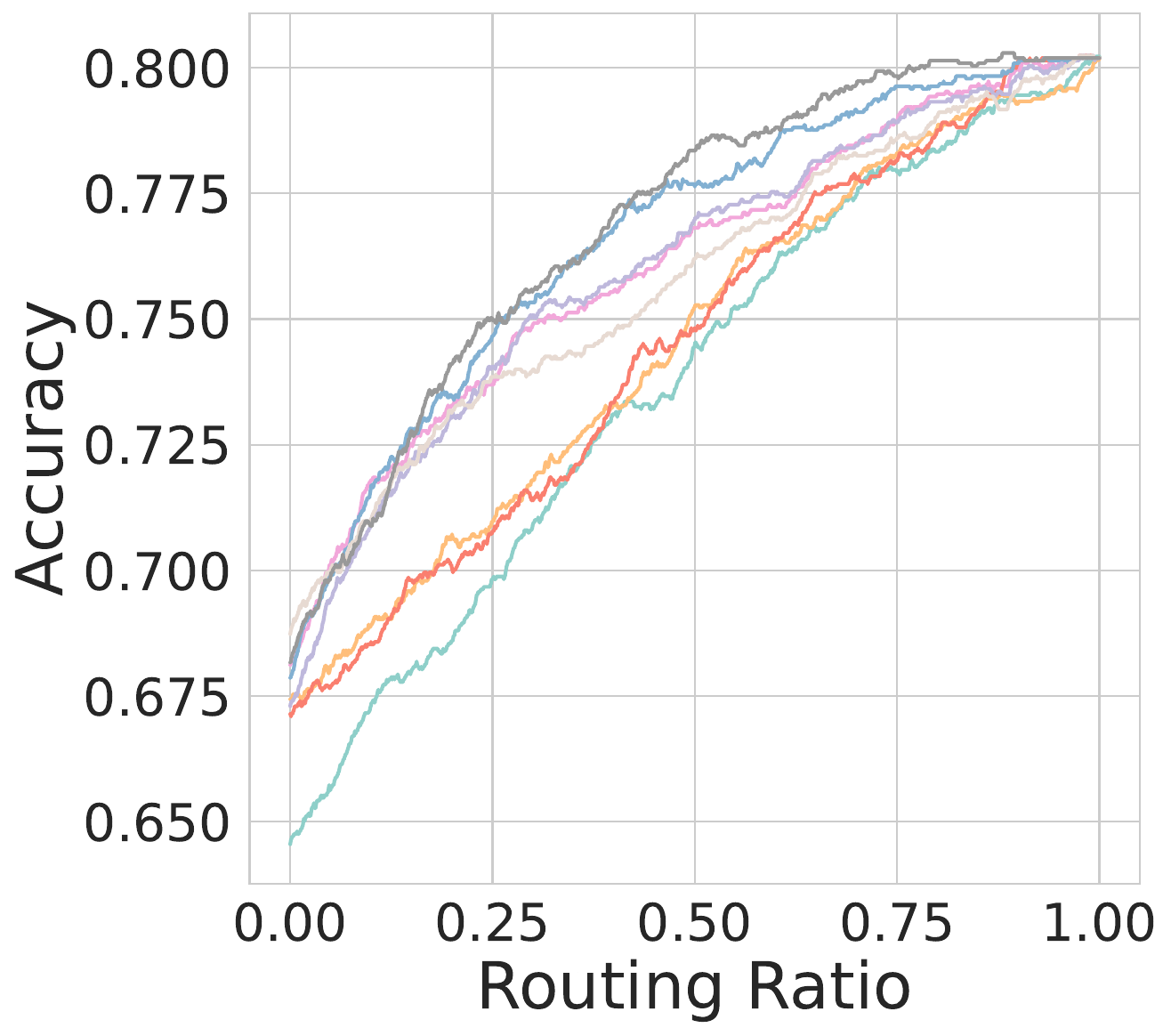}
    \end{minipage}%
    }
    \subfigure[Phi-3.5-mini]{
    \centering
    \begin{minipage}[t]{0.23\linewidth}
	    \includegraphics[width=1.05\linewidth]{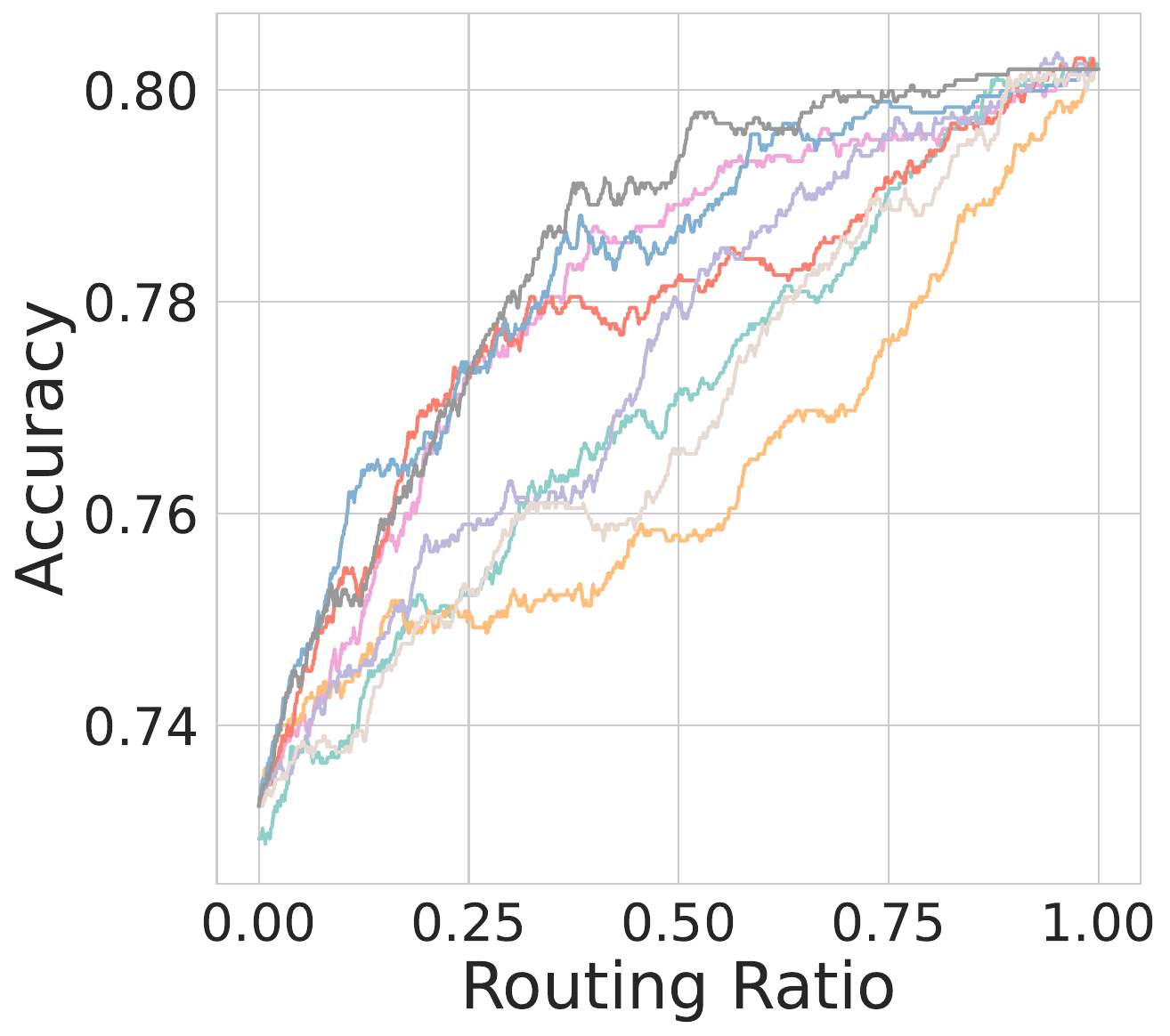}
    \end{minipage}%
    }
    \subfigure[Qwen-2.5-7B]{
    \centering
    \begin{minipage}[t]{0.23\linewidth}
	    \includegraphics[width=1.05\linewidth]{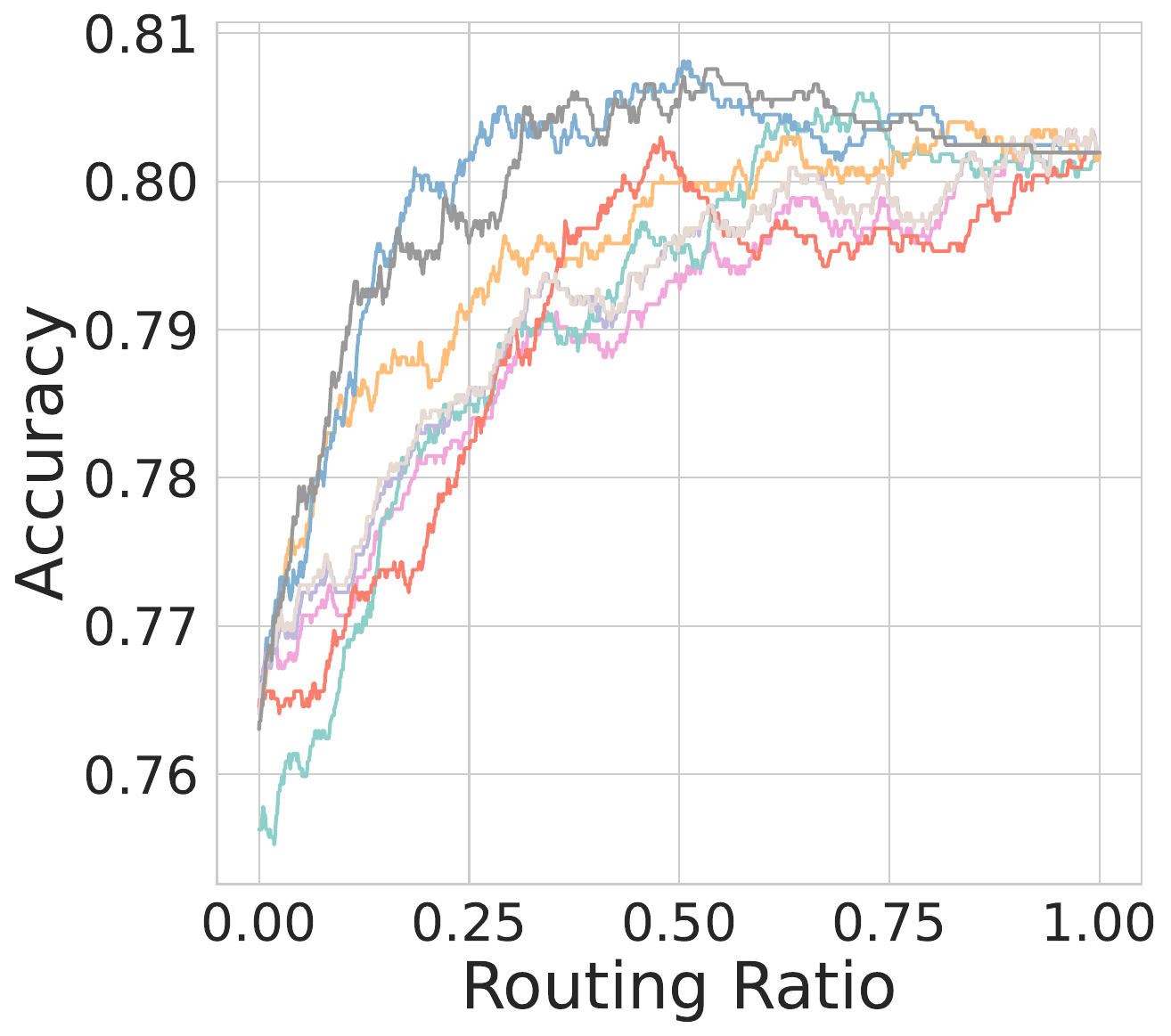}
    \end{minipage}%
    }
    \vspace{-2mm}
    \caption{Overall accuracy vs. routing ratio with different UQ methods and SLMs. (a)-(d) show the results of routing to Llama-3.1-70B on the OpenbookQA dataset; and (e)-(h) demonstrate the results of routing to GPT-4o mini on the Social IQa dataset.}
    \label{fig:routing_llama}
\end{figure*}

\subsection{Report Observations}
In this section, we present our benchmarking results analyzing the impact of uncertainty-correctness alignment on routing tasks. More observations and experimental results on calibration and routing can be found in Appendix~\ref{apdx:align}. 

\noindent\textbf{Observation \ding{182}: Uncertainty estimation in SLMs may exhibit misalignment with prediction correctness.} From the theoretical perspective, well-calibrated uncertainty scores do not necessarily imply a strong correlation with the correctness of the predictions~\cite{huang2023look,chuang2024learning}. The predictions of models might be perfectly calibrated yet still display relatively low accuracy (i.e., confidently provide wrong answers). This phenomenon is also evident in our benchmark results (illustrated in Figure~\ref{fig:uq_align}). We compute AUC scores to quantify the correlation between extracted uncertainty and prediction correctness, treating correctness as a binary ground truth and using confidence values as the ranking metric. The results show that not all UQ methods effectively exhibit a strong alignment between confidence and prediction correctness. Moreover, from Figure~\ref{fig:uq_align} and Figure~\ref{fig:uq_align_arith}, we can observe that the alignment may vary across datasets for the same SLM and UQ method. For instance, $p(\text{True})$~\cite{fadeeva2023lm, tian2023just} demonstrates strong alignment for Phi-3.5-mini on the MultiArith dataset but fails on the CommonsenseQA dataset. On the other hand, OOD Probe, Trained Probe, and Perplexity obtain consistently decent alignment compared to other UQ methods across different SLMs and domains of datasets. Conversely, we notice that verbalization-based methods, namely verbalization-1s~\cite{tian2023just, mahaut2024factual}, and verbalization-2s~\cite{tian2023just}, consistently withhold low alignment between uncertainty and prediction correctness. More experimental results can be found in Appendix~\ref{apdx:align}.

\noindent\textbf{Observation \ding{183}: Verbalization-based UQ methods struggle to extract uncertainty in SLMs for query routing.} 
We find that verbalization methods like verbalization-2s~\cite{tian2023just} obtain poor alignment between confidence and prediction correctness, and this misalignment can lead to inferior routing performance in SLMs, where the conclusion can be found in Figure~\ref{fig:routing_llama}. 
Recent advancements~\cite{xiong2023can,yona2024can} also show that uncertainty scores derived from verbalization may exhibit good reflection on models' intrinsic uncertainty of prediction across multiple models and datasets. This discrepancy poses a significant challenge for establishing effective routing performance since queries that are actually correct may be unnecessarily routed from SLMs to LLMs, thereby increasing the overall cost of deploying routing systems.

\begin{figure*}[!t]
    \centering
    \subfigure[Granite-3.1-8B]{
    \centering
    \begin{minipage}[t]{0.23\linewidth}
	    \includegraphics[width=1.0\linewidth]{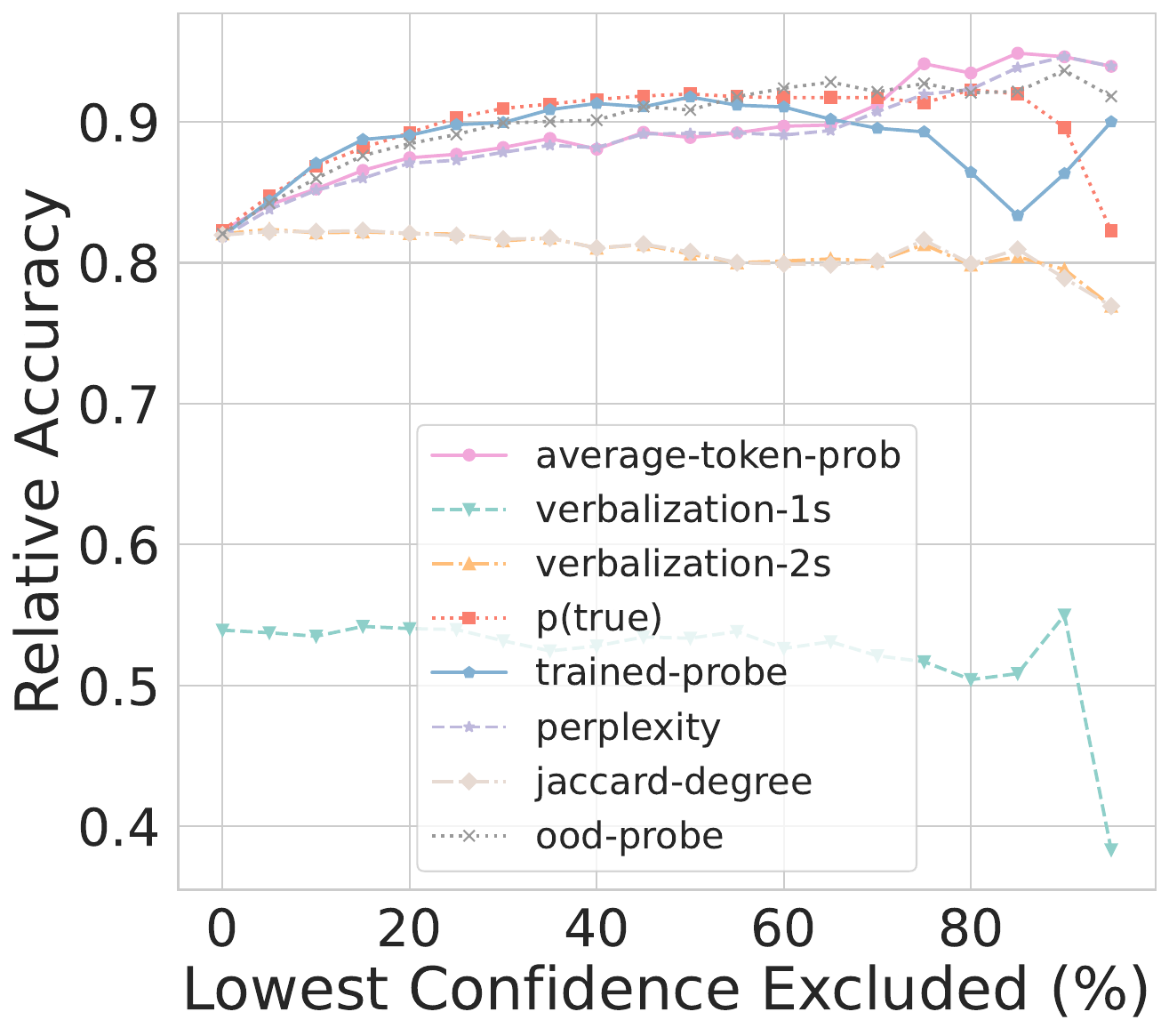}
    \end{minipage}%
    }
    \subfigure[Mistral-7B]{
    \centering
    \begin{minipage}[t]{0.23\linewidth}
	    \includegraphics[width=1.0\linewidth]{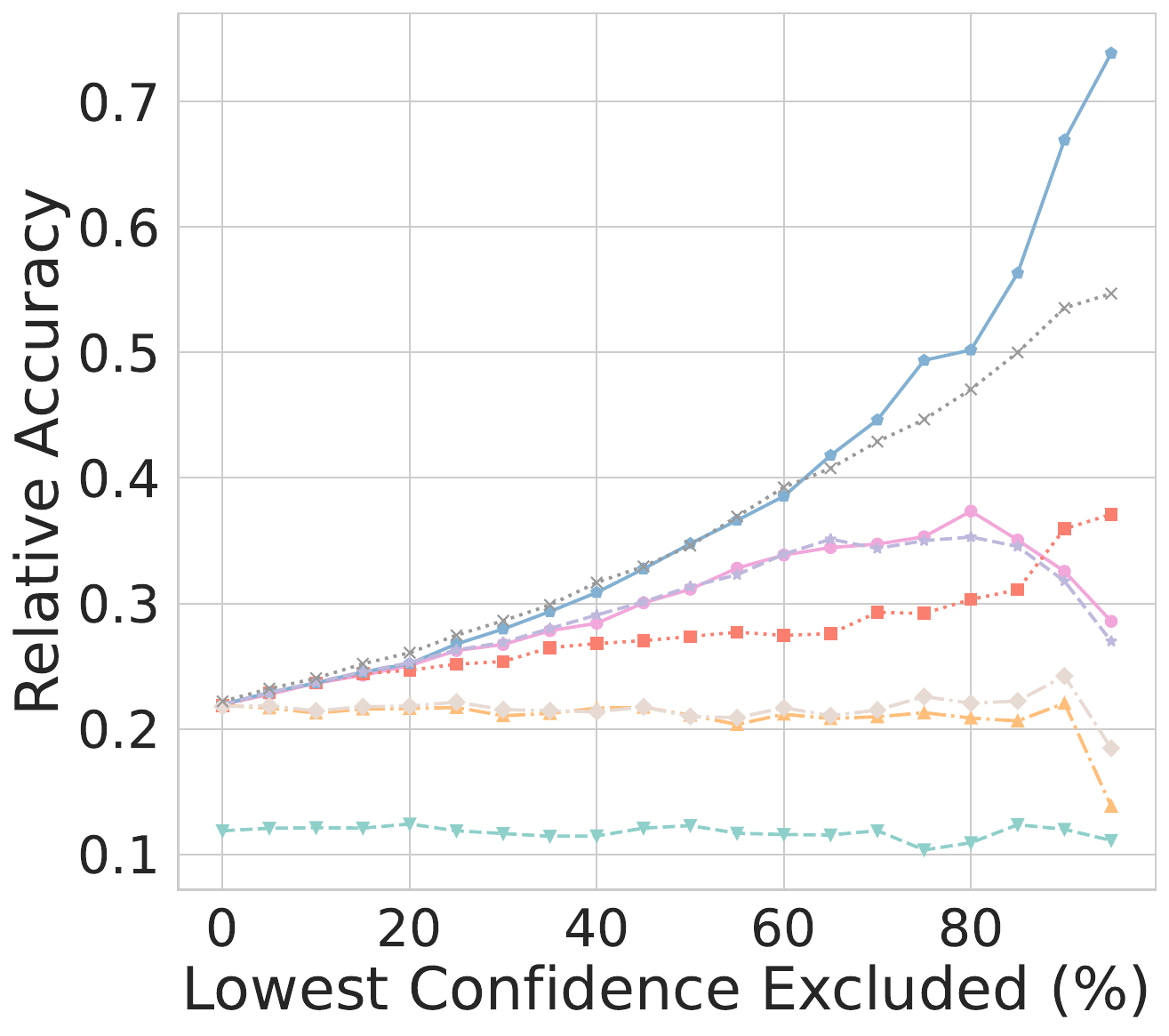}
    \end{minipage}%
    }
    \subfigure[Danube-3.1-4B]{
    \centering
    \begin{minipage}[t]{0.23\linewidth}
	    \includegraphics[width=1.0\linewidth]{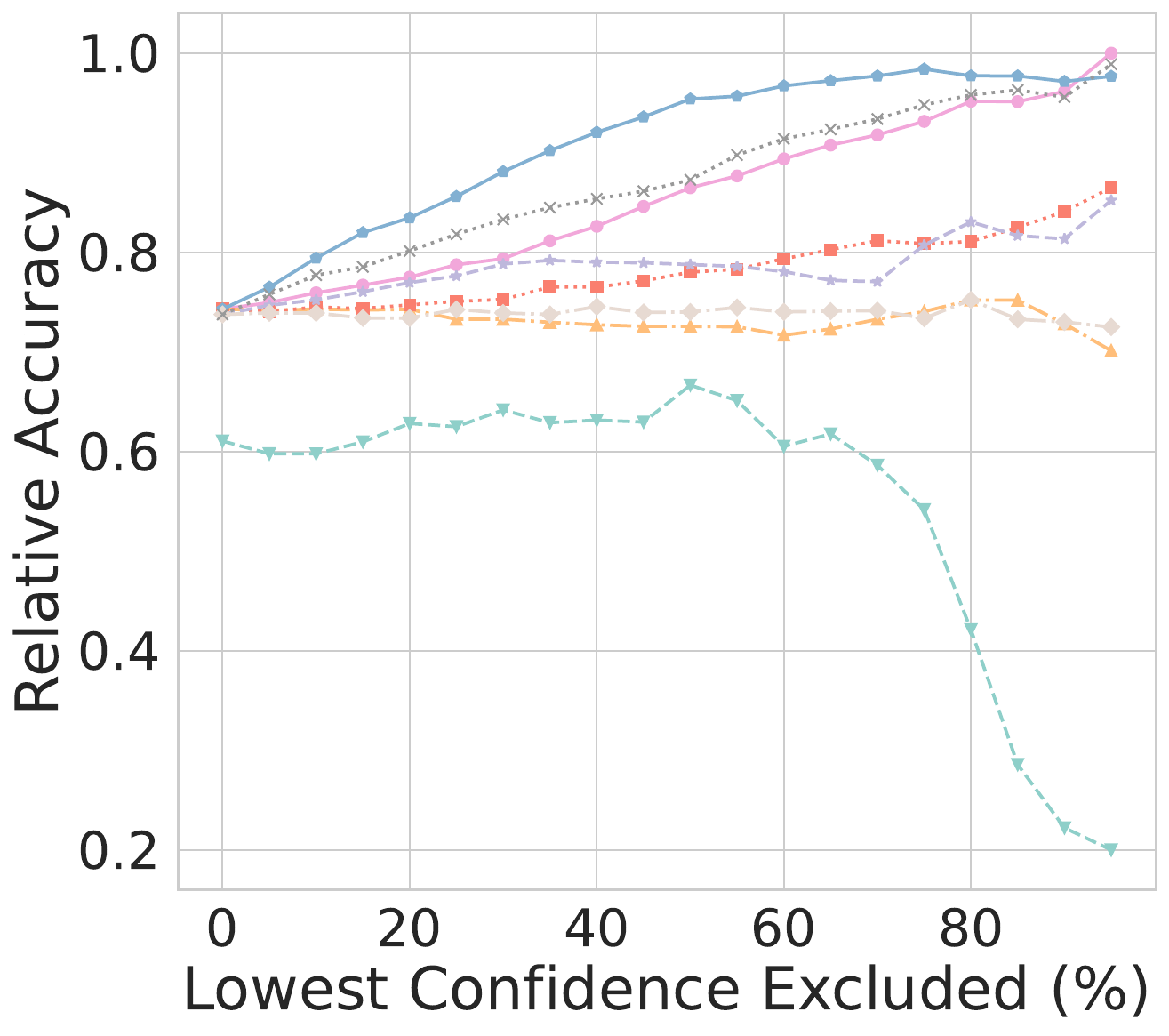}
    \end{minipage}%
    }
    \subfigure[Llama-3.1-8B]{
    \centering
    \begin{minipage}[t]{0.23\linewidth}
	    \includegraphics[width=1.0\linewidth]{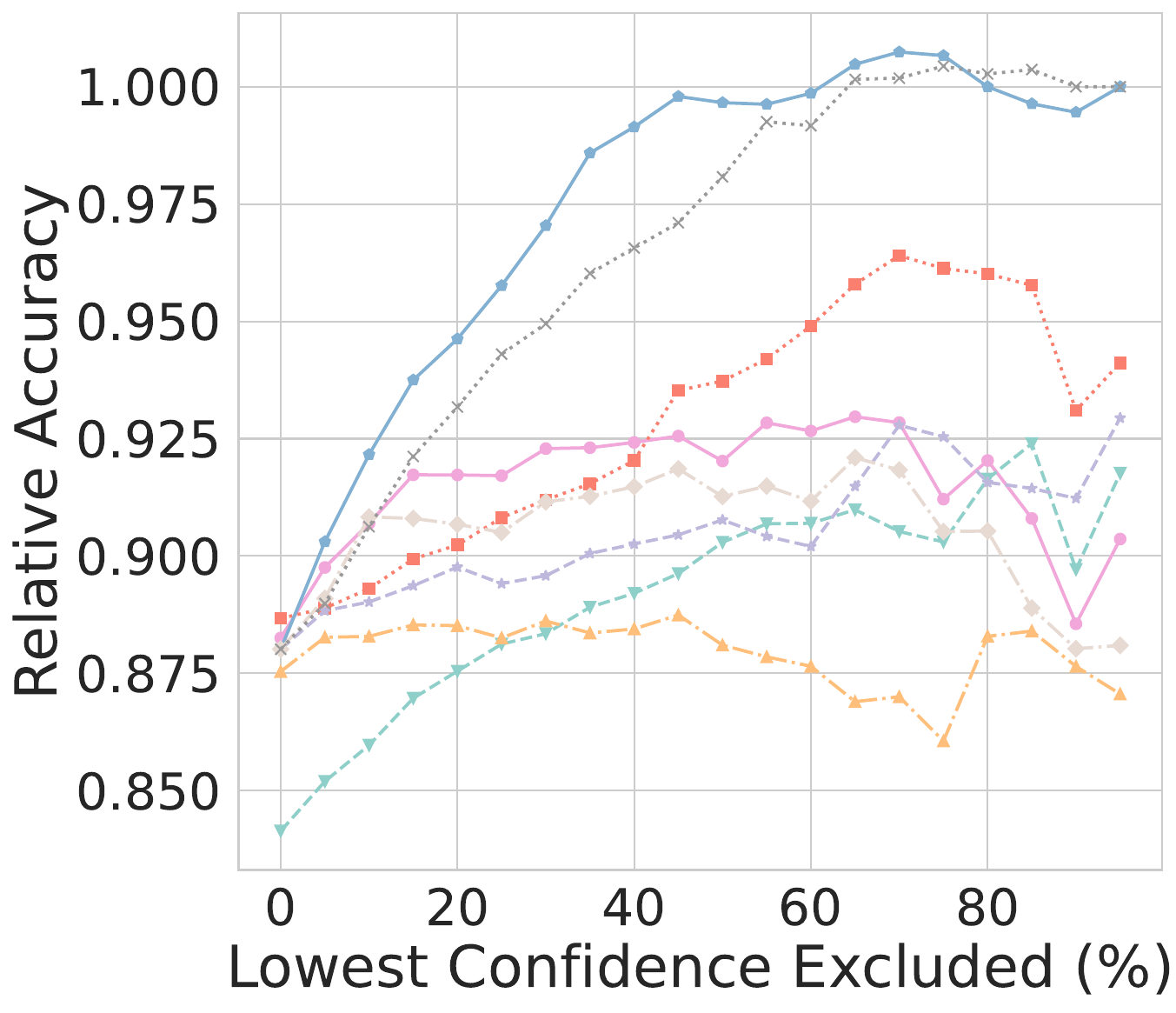}
    \end{minipage}%
    }
    \vspace{-4mm}
    \caption{Relative accuracy of SLMs vs. LLMs on top-$k\%$ confident queries. ``Relative accuracy'' is the ratio of SLM accuracy to LLM accuracy. 
    The x-axis ``Lowest Conf. Excluded'' shows the percentage of low-confidence queries removed; for example, 80 means 80\% of queries with the lowest confidence are excluded, leaving the top 20\%. 
    (a) and (b) compare SLMs to Llama-3.1-70B on GSM8K, while (c) and (d) compare SLMs to GPT-4o-mini on PIQA.}
    \label{fig:top_conf}
\end{figure*}

\noindent\textbf{Observation \ding{184}: A good routing standard highly depends on UQ methods with good uncertainty-correctness alignment.} A notable phenomenon occurs when UQ methods, such as Trained Probe~\cite{mahaut2024factual}, exhibit strong alignment, leading to significant improvements in routing performance. This is because the extracted uncertainty scores from these UQ methods more effectively indicate whether SLMs produce correct predictions.
Among all UQ methods evaluated for routing tasks, we find that Trained Probe~\cite{mahaut2024factual}, OOD Probe~\cite{kadavath2022language,mahaut2024factual}, and Perplexity~\cite{fadeeva2023lm} consistently rank as the top three methods for SLM routing. Therefore, a comprehensive analysis of UQ methods before deploying a routing system in SLMs is highly recommended to ensure efficient query routing.


\noindent\textbf{Observation \ding{185}: SLMs can match LLM performance on high-confidence queries.} Although SLMs generally underperform LLMs, we find that for queries where SLMs exhibit high confidence, their accuracy approaches that of LLMs. To illustrate, we progressively remove queries starting from those with the lowest SLM confidence and compute the ratio of SLM to LLM accuracy on the remaining top-$k\%$ queries (Figure~\ref{fig:top_conf}). As more low-confidence queries are excluded, SLMs achieve comparable performance to LLMs. For instance, on PIQA, Danube-3.1-4B achieves performance nearly equal to GPT-4o-mini on the top 20\% highest-confidence queries. Moreover, the effectiveness of this selection depends on the uncertainty quantification (UQ) method: approaches with stronger alignment (e.g., Trained Probe~\cite{mahaut2024factual}) yield higher relative accuracy than weaker ones (e.g., verbalization-2s) across all query exclusion rates. Additional results appear in Appendix~\ref{apdx:routing}.



\section{Generalizable SLM Routing for New Downstream Scenarios}
In this section, we first describe the pipeline for constructing calibration data with experimental details. We then investigate how well the calibration data can enhance routing generalization to new downstream scenarios without accessing the new datasets. 
Finally, we discuss our benchmark results and offer several insights into the calibration data for establishing routing in early-stage deployments.

\begin{figure*}[!t]
    \centering
    \subfigure[Phi-3.5-mini]{
    \centering
    \begin{minipage}[t]{0.46\linewidth}
	    \includegraphics[width=1.\linewidth]{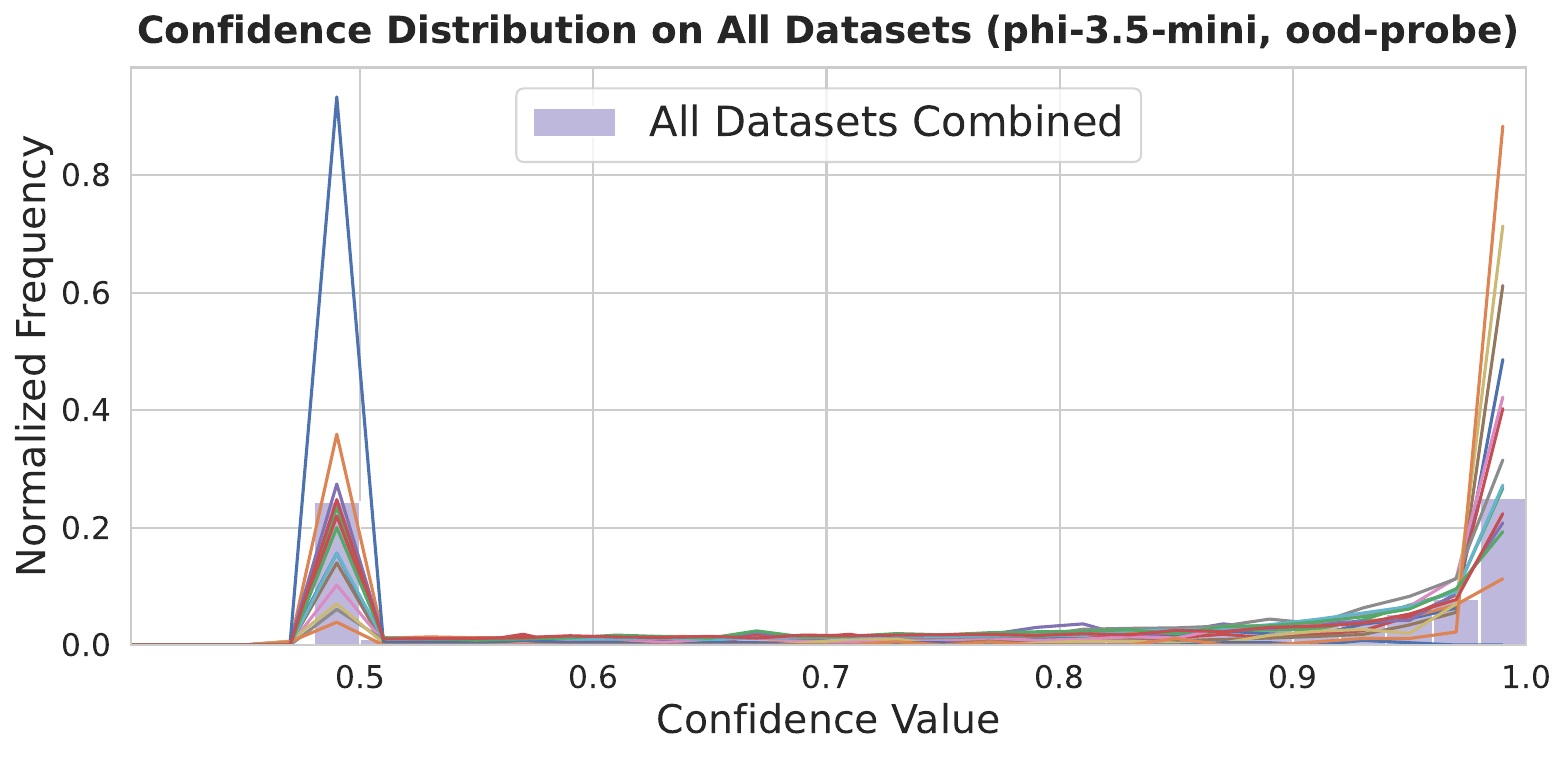}
    \end{minipage}%
    }
    \subfigure[Llama-3.1-8B]{
    \centering
    \begin{minipage}[t]{0.46\linewidth}
	    \includegraphics[width=1.\linewidth]{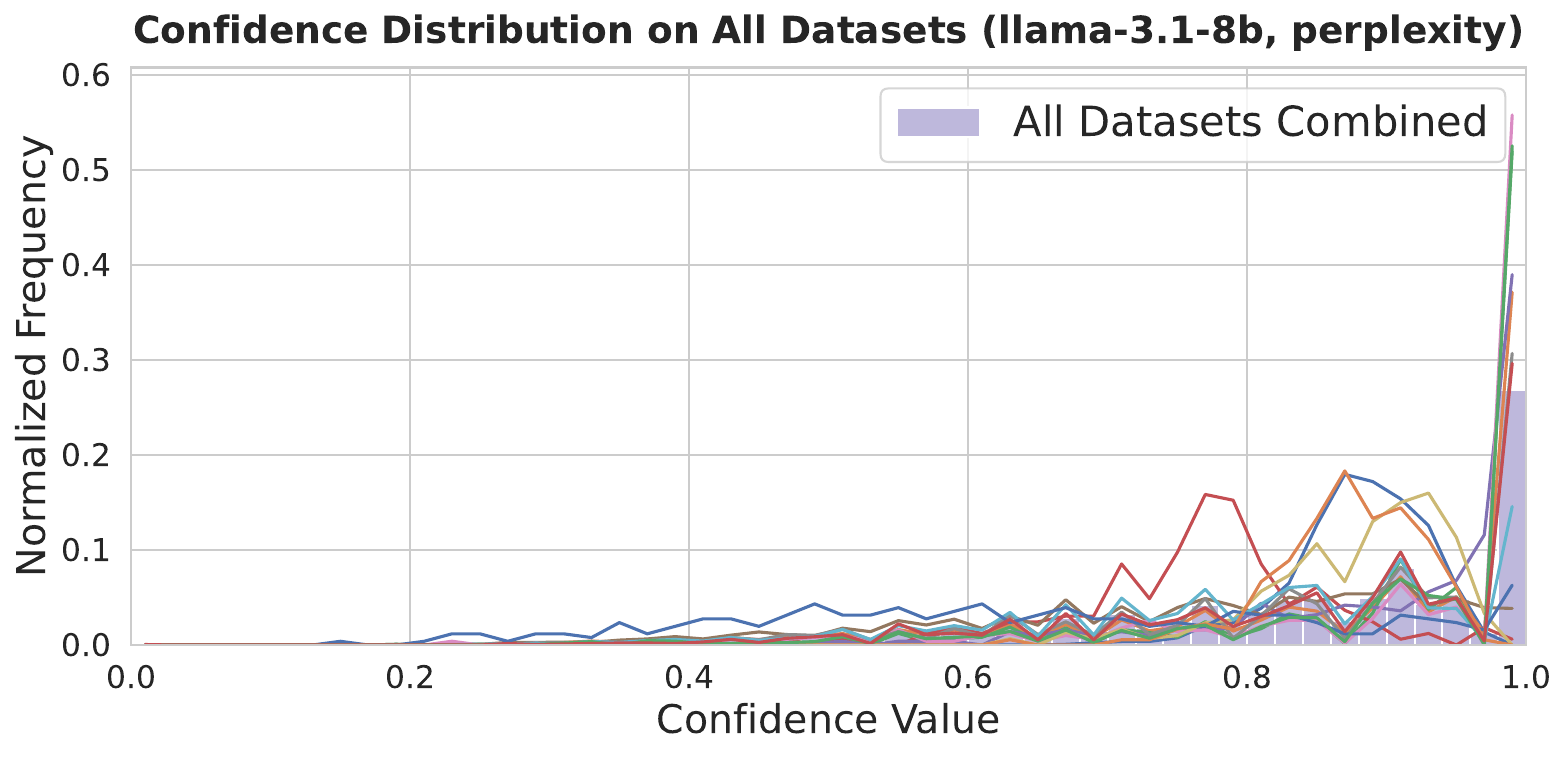}
    \end{minipage}%
    }
    \vspace{-2mm}
    \caption{Confidence distributions across $14$ datasets. The histogram depicts the aggregated distribution from all datasets, while each curve represents a single dataset. (a) Confidence of Phi-3.5-mini by OOD Probe; (b) Confidence of Llama-3.1-8B by Perplexity.}
    \label{fig:cali_dist}
\end{figure*}

\begin{figure*}[!t]
    \centering
    \includegraphics[width=1.\linewidth]{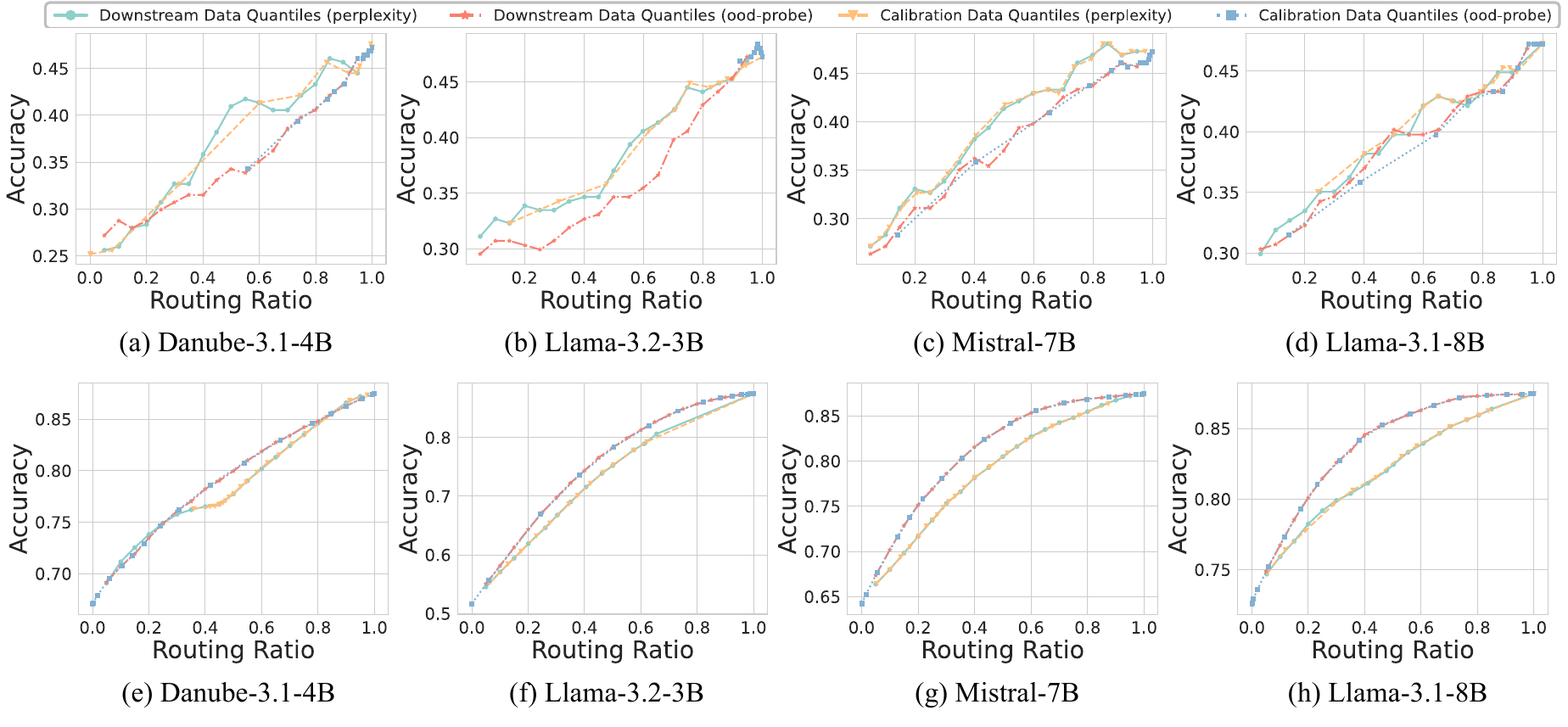}
    \vspace{-8mm}
    \caption{Generalization of calibration data for routing on new downstream datasets. (a)-(d) show the results of routing to Llama-3.1-70B on AQuA; (e)-(h) show results of routing to GPT-4o-mini on HellaSwag.}
    \label{fig:cali_gen}
    \vspace{-2mm}
\end{figure*}

\subsection{Calibration Data Construction Pipeline}
\label{sec:cali_pip}
We aim to evaluate how calibration data improves routing strategies when generalizing to new downstream scenarios without requiring additional downstream data. Specifically, the calibration data serves as a data-agnostic hold-out set tailored to a particular SLM, which can generalize its routing standards across various new downstream datasets. 
By leveraging this calibration data, we establish a generalizable routing framework for new scenario deployments. This approach not only simplifies the deployment process by removing the need for comprehensive dataset-specific analytics for routing but also demonstrates the potential of calibration data as a proxy for diverse data types, ultimately enhancing the generalization to new scenarios.

\begin{algorithm}[t!]\small
\caption{Calibration Data Construction Pipeline}
\label{alg:data_aug}
\begin{algorithmic}[1]
\INPUT A collection of datasets $\mathbb{D} = \{\mathcal{D}_i\}_{i=1}^{N}$ with $N$ domains
\OUTPUT A set of calibration data $\boldsymbol{X}$ 

\STATE Collect diverse domain of dataset $\mathcal{D}_i$ to form $\mathbb{D} = \{\mathcal{D}_i\}_{i=1}^{N}$
\STATE Generate uncertainty distributions $\{\mathcal{F}_{\mathbb{D}}\}_{i=1}^{M}$ of $\mathbb{D}$ with selected UQ methods
\STATE Sample $\boldsymbol{X} = \{ x_j ~|~  x_j \in \boldsymbol{X}_i \sim \{\mathcal{F}_{\mathbb{D}}\}_{i=1}^{M} ~\forall i, j\}$ from $i$-th bin in $\{\mathcal{F}_{\mathbb{D}}\}_{i=1}^{M}$
\end{algorithmic}
\end{algorithm}
\raggedbottom

The overall construction pipeline is detailed as follows. Let $\mathbb{D} = \{\mathcal{D}_i\}_{i=1}^{N}$ be a diverse collection of datasets, where $N$ denotes the number of distinct domain types included in the collection. We select a diverse collection of datasets $\mathbb{D}$ with various domains, such as commonsense reasoning, mathematics, and more, where we follow the settings in \cite{liu2024dora}. 
And then, we process every data instance in $\mathbb{D}$ through selected UQ methods to capture their corresponding uncertainty distributions $\{\mathcal{F}_{\mathbb{D}}\}_{i=1}^{M}$ with $M$ bins, where $M \in \mathbb{Z}^{+}$ is an arbitrary number. These distributions serve as the sampling foundation of each data instance in forming calibration data.
Finally, data instances obtained in the set of calibration data $\boldsymbol{X}$ are weighted-sampled from each bin of $\{\mathcal{F}_{\mathbb{D}}\}_{i=1}^{M}$ such that $\boldsymbol{X} = \{ x_j ~|~  x_j \in \boldsymbol{X}_i \sim \{\mathcal{F}_{\mathbb{D}}\}_{i=1}^{M} ~\forall i, j\}$. This ensures similar distribution in calibration data across various uncertainty levels presented in $\{\mathcal{F}_{\mathbb{D}}\}_{i=1}^{M}$. The resulting collection of these sampled data instances forms the final calibration dataset. The detailed pipeline of constructing the calibration dataset is outlined in Algorithm~\ref{alg:data_aug}.

\subsection{Calibration Data Setups}

\noindent\textbf{Benchmark Settings.}
We evaluate the constructed calibration data on $8$ SLMs and $2$ LLMs across $14$ datasets. Based on the observations and results from the previous benchmark section, we select $2$ UQ methods that demonstrate the strongest alignment between predicted uncertainty and actual correctness: "OOD Probe"~\cite{kadavath2022language, mahaut2024factual} and "Perplexity"~\cite{fadeeva2023lm} method. We consider the routing performance evaluated on entire new datasets as the ground truth.
To simulate a new dataset scenario, we adopt a cross-validation setting. Specifically, we select a target dataset for evaluation and derive its calibration data from the remaining $13$ datasets. The target dataset’s generalization performance is then evaluated using this calibration set, which does not contain any information from the target dataset. All reported results represent the average across three individual experimental runs. 

\noindent\textbf{Data Construction Settings.} The calibration data is weighted-sampled from each bin of the calibration data distributions, with the number of bins set to $30$. We sample $10\%$ of the instances from each bin to form the final calibration data. The temperature is fixed at $0$, top-p at $1.0$, and all decoding processes use greedy search. Alternative sampling strategies are disabled with a fixed random seed of $50$ to ensure reproducibility.

\subsection{Generalization of Calibration Data}
We provide several key insights into the generalization ability of calibration data as follows.

\noindent\textbf{Insights \ding{182}: The extracted confidence distribution is predominantly determined by the chosen SLM and uncertainty quantification (UQ) method, with minimal dependence on the downstream dataset.} As illustrated in Figure~\ref{fig:cali_dist}, confidence scores aggregated from $14$ different tasks exhibit a nearly identical shape regardless of the specific dataset. Instead, they vary notably with different SLMs and UQ methods. This finding suggests that the confidence distribution is largely data-agnostic, enabling the construction of calibration data that generalizes to new tasks without needing prior knowledge of new datasets.

\noindent\textbf{Insights \ding{183}: Calibration data helps SLM routing to achieve strong generalization, allowing routing strategies to be initialized on SLMs without accessing new datasets.} Building on our findings about uncertainty distributions, we sampled a data subset to create a final calibration dataset using the pipeline described in Section~\ref{sec:cali_pip}. We then utilized this calibration dataset to determine thresholds for different routing ratios in new downstream scenarios. The experimental results (see Figure~\ref{fig:cali_gen}) show that the routing curves from the calibration data closely match those from the entire new downstream dataset, indicating that the calibration data provides strong generalization for establishing routing strategies on unseen downstream datasets. An identical phenomenon is observed across multiple UQ methods and different SLMs, highlighting the potential of calibration data to initiate the routing process for any new dataset independent of the UQ method or SLM used. Additional experimental results can be found in Appendix~\ref{apdx:generalization}.

\section{Challenges and Opportunities}

\noindent\textbf{\ding{182} How to cash-in routing efficiency on new edge devices?} Based on the benchmark results, calibration data provides a robust foundation for establishing routing policies on new edge devices without accessing prior knowledge at the early stage of deployment. This enables the routing policies with strong generalization to new dataset scenarios and enhances the efficiency across diverse deployment for personal edge devices. While calibration data holds a good performance in the early deployment stage, an important direction to explore is how to effectively leverage additional private on-device data to further strengthen the quality of calibration data, aiming to continuously enhance the deployment of personalized routing strategies. With the aid of calibration data, less private data is required, but striking a balance between privacy and performance remains an open challenge.

\noindent\textbf{\ding{183} How to effectively strike a balance between LLM routing efficiency and utility?} We empirically observe that by leveraging UQ methods with strong uncertainty-utility alignment (e.g., Perplexity and OOD Probe methods), routing thresholds can effectively be determined with the sweet points of efficiency and utility. However, achieving such sweet spots can be challenging due to the variability in downstream datasets and the sensitivity of UQ methods to LLM-specific characteristics. Additionally, discrepancies across different device types, such as variations between iOS and Android systems\footnote{iOS uses Apple-specific hardware, while Android spans a wide variety of ARM-based processors with different capabilities.}, further complicating the process, requiring tailored strategies and analytics to account for platform-specific constraints and capabilities. Based on these factors, providing a fair apple-to-apple comparison regarding routing performance is inherently challenging. Researchers should be mindful of these complexities and focus on developing methods that are not only efficient but also capable of handling long-context scenarios effectively.

\noindent\textbf{\ding{184} How is the performance when conducting compression (e.g., pruning, quantization) on the on-device model?}
As with the on-device models discussed in the above sections, we directly adopt a pre-trained small model without any modifications. Alternatively, on-device models can also be generated by compressing larger models. Specifically, numerous works have explored methods for compressing LLMs into smaller sizes using techniques such as pruning \cite{frantar2023sparsegpt}  and quantization \cite{xiao2023smoothquant, lin2024awq}. The advantage of employing compression methods is that the smaller models compressed from larger ones tend to retain similar distributions of the output, thereby mitigating the issue of distribution shift. Therefore, directly applying compression techniques to larger models is a promising field to explore in the future.

\noindent\textbf{\ding{185} Uncertainty-aware routing in on-device multimodal language models.} 
While LLMs typically operate with a single modality for both input and output, a promising research direction involves exploring uncertainty-aware routing in multimodal language models (MLLMs). For instance, in vision-language models (VLMs) such as LLaVa \cite{liu2024visual} and InternVL \cite{chen2024internvl}, the inputs include both images/videos and text. By incorporating visual modalities, the properties of vision tokens significantly influence the output. As a result, the uncertainty in the generated text differs from that of language-only models. Benchmarking and generalizing uncertainty-aware routing for on-device MLLMs is a valuable direction for the research community.

\section{Conclusion}
This paper investigates the routing accuracy of SLMs in estimating their uncertainty and establishing best practices for initiating effective routing strategies. Through comprehensive benchmarking of $8$ SLMs, $2$ LLMs, $8$ UQ methods, and $14$ datasets across 1500+ settings, we found that the alignment between uncertainty and correctness significantly impacts routing performance. Additionally, our experiments show that uncertainty distributions depend primarily on the specific SLM and UQ method rather than the downstream data. Building on the insights, we introduced a calibration data construction pipeline and a hold-out dataset to generalize routing strategies without prior knowledge of new downstream data. The results confirm that the calibration data effectively bootstraps routing, indicating its strong potential for benefiting in resource-efficient SLM deployment.


\bibliography{main}
\bibliographystyle{icml2021}

\clearpage
\appendix
\onecolumn

\section*{Appendix}

\section{Details about Datasets} \label{app:data}
The details of the $14$ datasets are further listed in Tabel~\ref{tab:datasets}. We applied the original dataset directly from the Huggingface dataset repositories without any further processing. A thorough examination of each dataset’s attributes, size, and notable characteristics is provided below.

\begin{table*}[h!]
    \centering
    \caption{Details of the $14$ datasets used in our benchmark. FF: Free-form question answering (including numerical answers for math tasks); MCQ: Multiple-choice question answering; TF: True/False question answering.}
    \vspace{0.25cm}
    \resizebox{1.\textwidth}{!}{%
    \begin{tabular}{lccccc}
        \toprule
        Dataset & Type & Domain & \# Train & \# Test & Description\\
        \midrule
        GSM8K & FF & Mathematical Reasoning & $7473$ & $1319$ & Grade school math word problems\\
        AQuA & MCQ & Mathematical Reasoning & $97467$ & $254$ & Algebraic word problems\\
        MultiArith & FF & Mathematical Reasoning & $420$ & $180$ & Algebraic word problems\\
        SVAMP & FF & Mathematical Reasoning & $700$ & $300$ & Algebraic word problems\\
        \midrule
        BoolQ & TF & Commonsense Reasoning & $9427$ & $3270$ & Commonsense and factual reasoning questions\\
        CommonsenseQA & MCQ & Commonsense Reasoning & $9741$ & $1221$ & Questions assessing various types of commonsense knowledge\\
        HellaSwag & MCQ & Commonsense Reasoning & $39905$ & $10042$ & Sentence completion based on narrative understanding\\
        OpenBookQA & MCQ & Commonsense Reasoning & $4957$ & $500$ & Open-book science and commonsense questions\\
        PIQA & MCQ & Commonsense Reasoning & $16113$ & $1838$ & Physical commonsense reasoning questions\\
        Social IQa & MCQ & Commonsense Reasoning & $33410$ & $1954$ & Social commonsense intelligence questions\\
        TruthfulQA & FF & Commonsense Reasoning & $653$ & $164$ & Assessing models' ability to prevent false information\\
        WinoGrande & MCQ & Commonsense Reasoning & $2558$ & $1267$ & Pronoun ambiguity resolution with commonsense reasoning\\
        \midrule
        CoQA & FF & Conversational \& Contextual Understanding & $7199$ & $500$ & Conversational questions on text passages from diverse domains\\
        \midrule
        MMLU & MCQ & Problem Solving & $99842$ & $14042$ & Problem solving across various subjects\\
        \bottomrule
    \end{tabular}}
    \label{tab:datasets}
\end{table*}

\section{Related Work}

\subsection{Small Language Models} Small Language Models (SLMs) are designed for deployment on resource-constrained devices like desktops, smartphones, and wearables.
Specifically, we consider the Transformer-based  SLMs in this work due to their state-of-the-art performance, like Phi-3-mini~\cite{abdin2024phi}, TinyLlama~\cite{zhang2024tinyllama}, MobileLLM~\cite{liu2024mobilellm}, and Qwen-1.5B~\cite{bai2023qwen}, LiteLLaMa-460M, OPT-125M~\cite{zhang2022opt}, BLOOMZ~(560M, 1.1B, 1.7B, 3B)~\cite{le2023bloom}, SmolLM~(135M, 360M, 1.7B)~\cite{allal2024SmolLM}, OLMo~(1B)~\cite{Groeneveld2023OLMo}, OLMoE~(1B)~\cite{muennighoff2024olmoeopenmixtureofexpertslanguage}, MobiLlama~(0.5B, 1B)~\cite{thawakar2024mobillama}, MobileLLaMA~(1.4B, 2.7B)~\cite{chu2024mobilevlm}, OpenLLaMA~(3B)~\cite{openlm2023openllama}.
These models are designed with lightweight architectures to operate effectively within the computational and storage limitations of mobile devices and edge hardware.

Recurrent Neural Networks (RNNs), like RWKV~(1B, 3B, 7B)~\cite{peng2023rwkv}, Mamba~(1.4B, 6.9B)~\cite{dao2024transformers}, and RecurrentGemma-2B~\cite{griffin2024recurrentgemma}, can provide promising solutions for on-device inference in resource-constrained environments. 
These models leverage the recurrent nature of RNNs to process sequential data efficiently without requiring a KV cache, which is suitable for resource-constrained on edge devices. 
Specifically, RWKV introduces a hybrid RNN-Transformer backbone to capture long-term dependencies while maintaining computational efficiency. 
Similarly, Mamba and RecurrentGemma design recurrent layers for low-power consumption and high throughput inference, which can significantly reduce memory and computational requirements, fostering low-latency applications directly on devices.

\clearpage
\section{Additional Experimental Results from Benchmarking to Generalization }
\label{apdx:add_exp}

In this section, we present additional experimental results on (1) evaluating the impact of uncertainty-correctness alignment on small language model (SLM) routing and (2) investigating the generalization capability of calibration data on novel datasets.
\textit{For the first experiment (Section~\ref{apdx:align} and Section~\ref{apdx:routing})}, we provide the complete set of results, including the AUC measurements for uncertainty-correctness alignment and the performance of uncertainty-based routing.
\textit{For the second experiment (Section~\ref{apdx:generalization})}, we present a comprehensive analysis of the generalization ability of calibration data to unseen downstream datasets. Each dataset referenced in the experiment is treated as a novel dataset for evaluation. We adopt a cross-validation setup, ensuring that the calibration data does not include any instances from the target dataset.

\subsection{Evaluation on Uncertainty-correctness Alignment} \label{apdx:align}
\subsubsection*{Results of Alignment between uncertainty and correctness.}
All the experiments shown on this page are conducted under AQUA, BoolQ, and CoQA datasets with all $8$ UQ methods.

\begin{figure*}[h!]
\centering
    \includegraphics[width=0.9\textwidth]{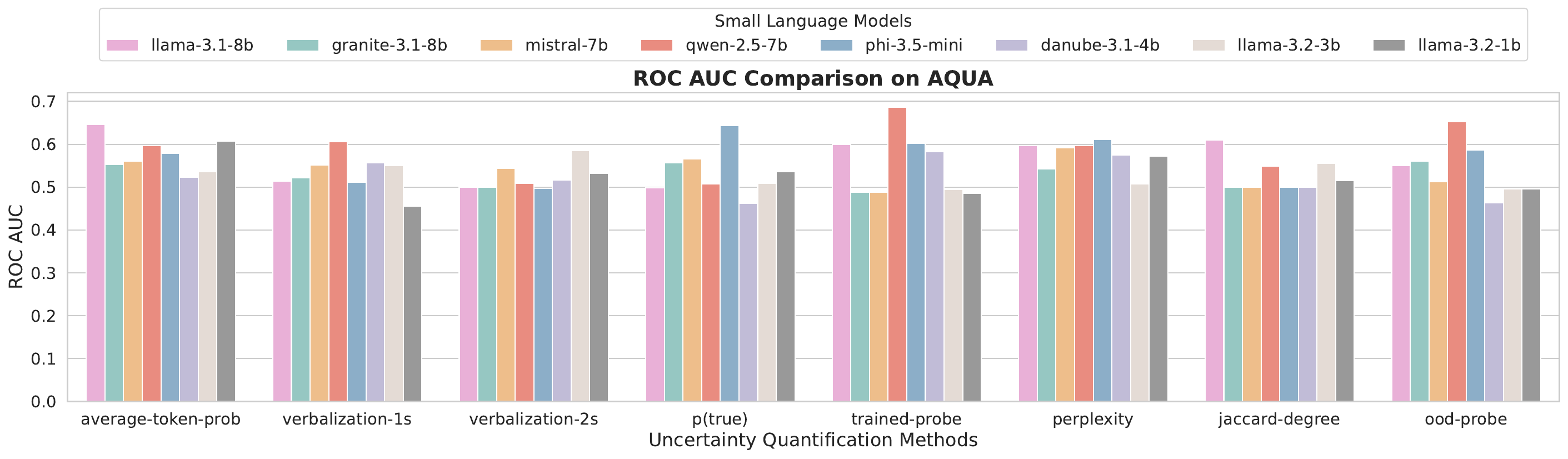} 
    \includegraphics[width=0.9\textwidth]{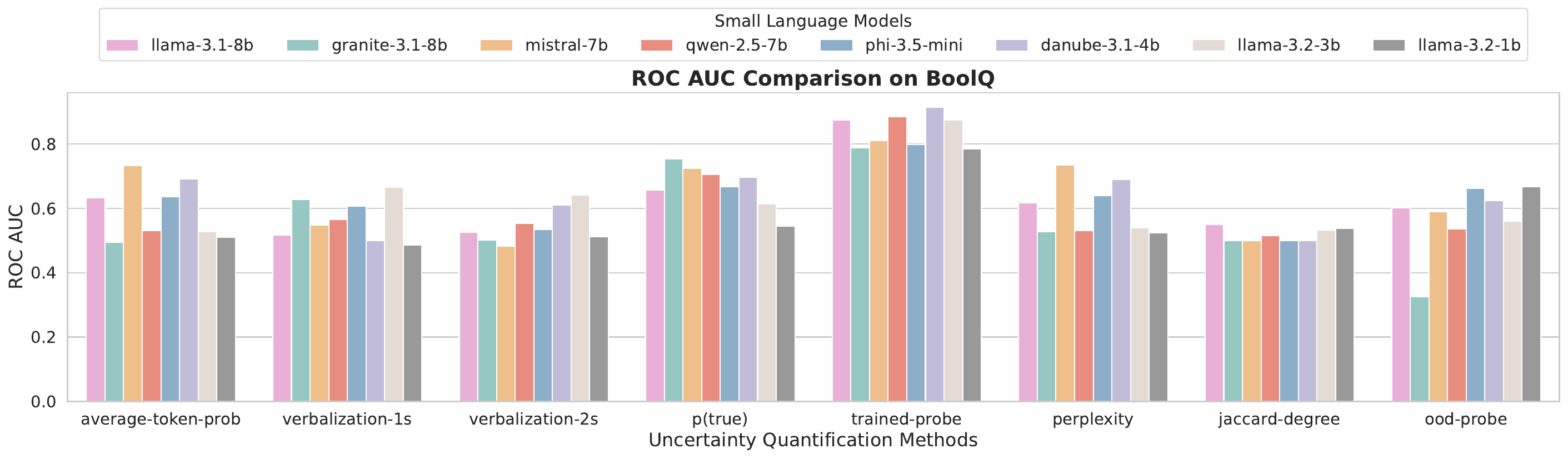}
    \includegraphics[width=0.9\textwidth]{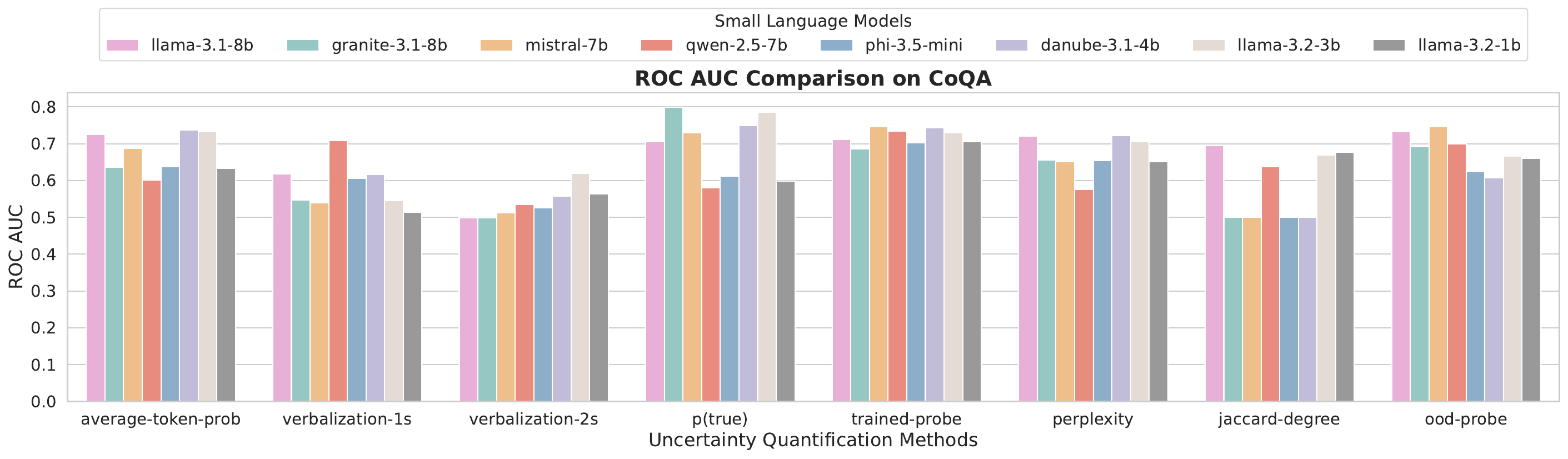}
    \vspace{-3mm}
    \caption{The ROC AUC scores measure the alignment between confidence and correctness across different SLMs and uncertainty quantification methods on AQuA, BoolQ, and CoQA. A higher ROC AUC indicates a stronger alignment.}
\label{fig:uq_align1} 
\end{figure*}

\clearpage
\subsubsection*{Results of Alignment between uncertainty and correctness.}
All the experiments shown on this page are conducted under GSM8K, HellaSwag, MMLU, and MultiArith datasets with all $8$ UQ methods.

\begin{figure*}[h!]
\centering
    \includegraphics[width=0.9\textwidth]{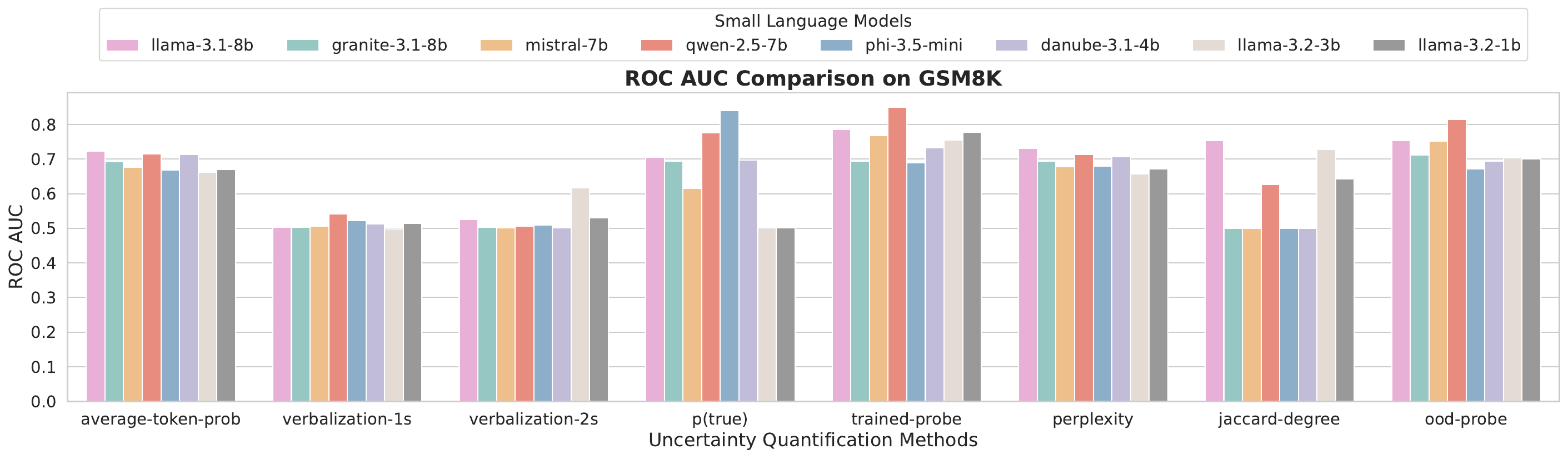} 
    \includegraphics[width=0.9\textwidth]{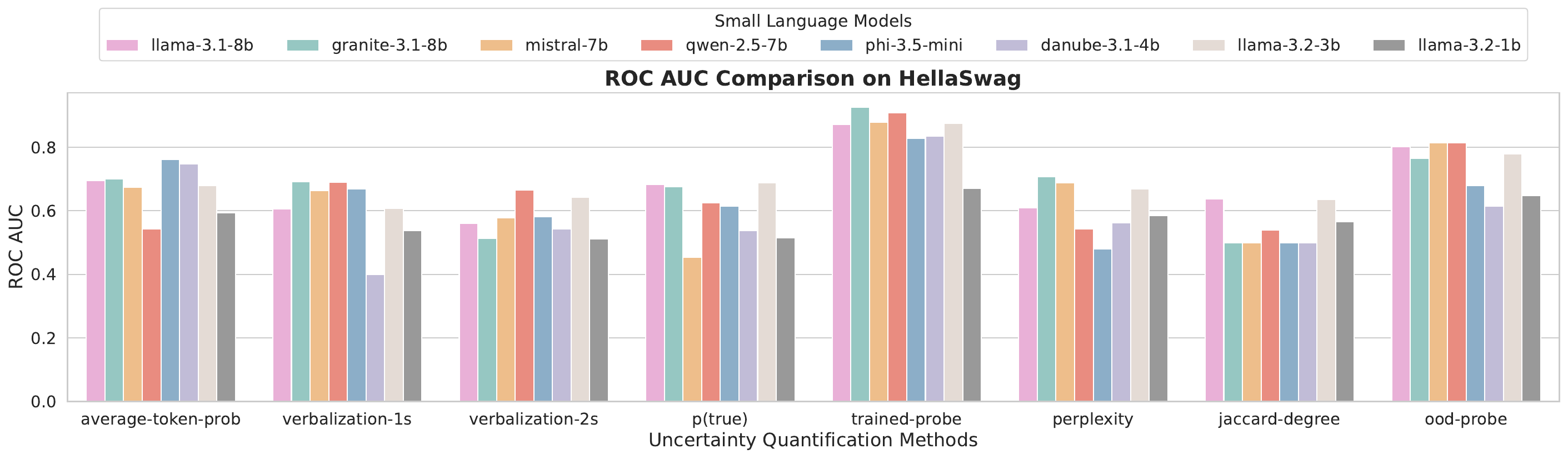}
    \includegraphics[width=0.9\textwidth]{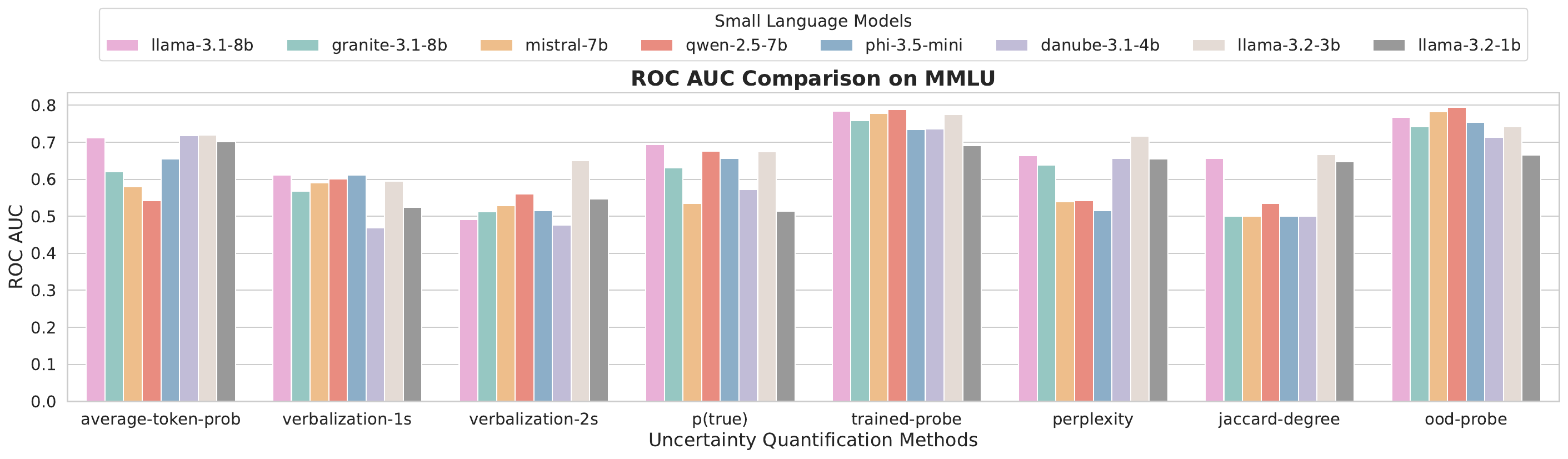}
    \includegraphics[width=0.9\textwidth]{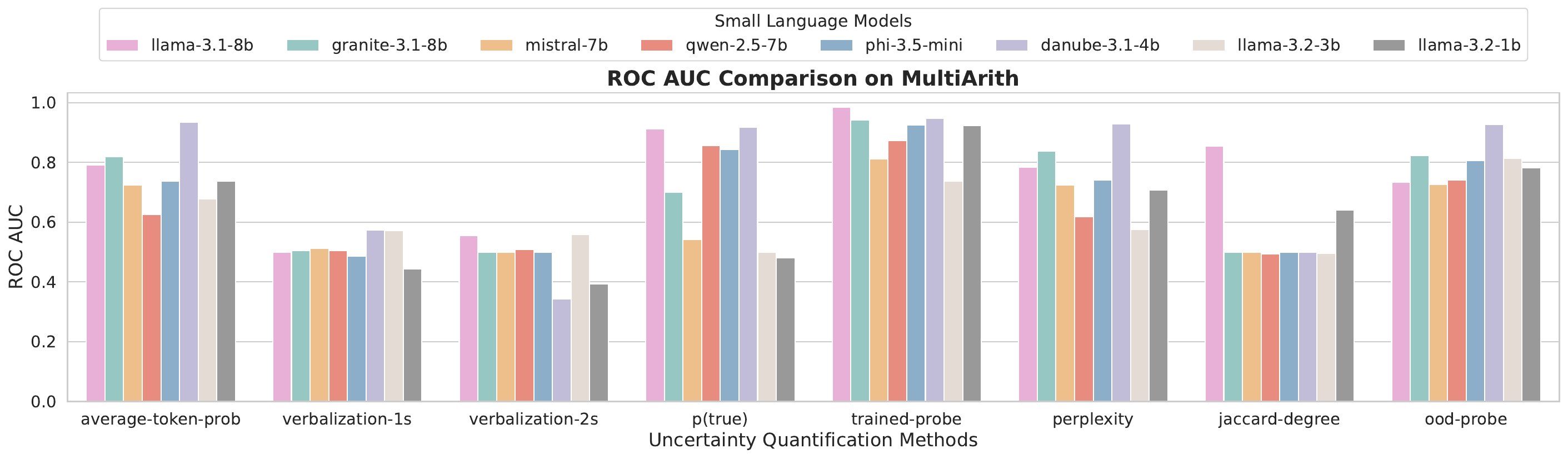}
    \vspace{-3mm}
    \caption{The ROC AUC scores measure the alignment between confidence and correctness across different SLMs and uncertainty quantification methods on GSM8K, HellaSwag, MMLU, and MultiArith. A higher ROC AUC indicates a stronger alignment.}
\label{fig:uq_align_arith} 
\end{figure*}

\clearpage
\subsubsection*{Results of Alignment between uncertainty and correctness.}
All the experiments shown on this page are conducted under OpenBookQA, PIQA, SocialIQA, and SVAMP datasets with all $8$ UQ methods.

\begin{figure*}[h!]
\centering
     \includegraphics[width=0.9\textwidth]{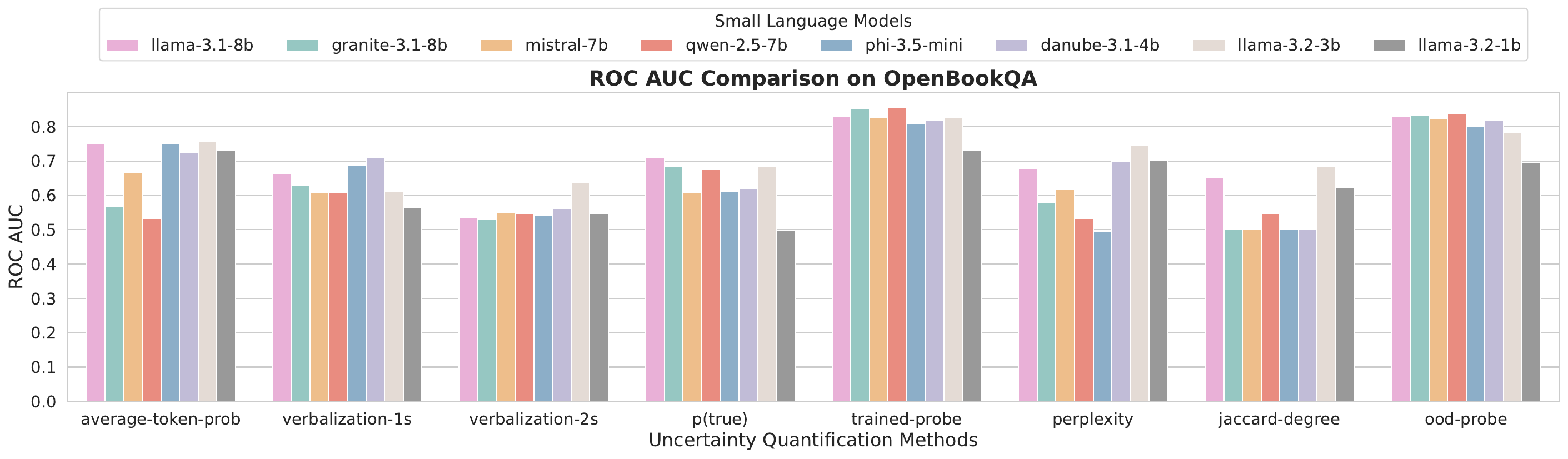}
    \includegraphics[width=0.9\textwidth]{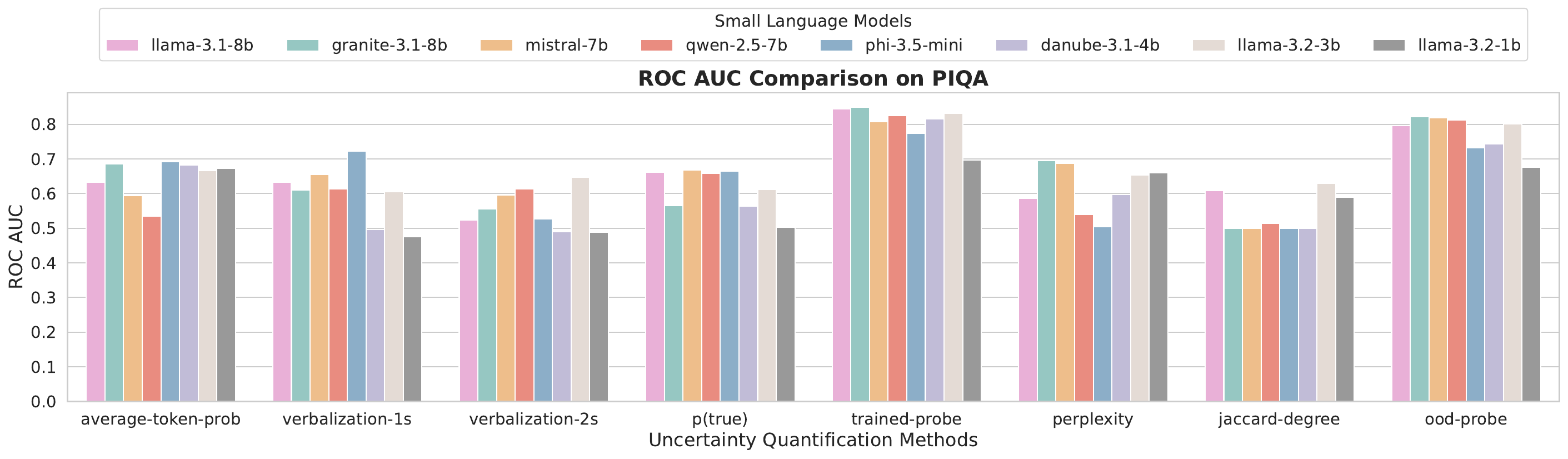} 
    \includegraphics[width=0.9\textwidth]{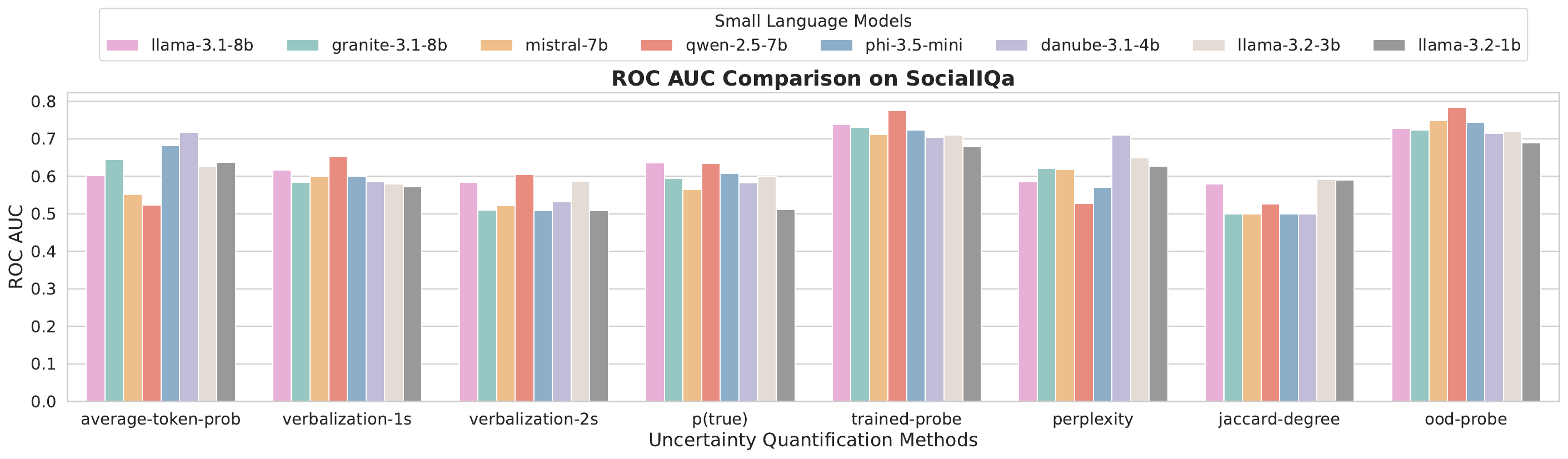}
    \includegraphics[width=0.9\textwidth]{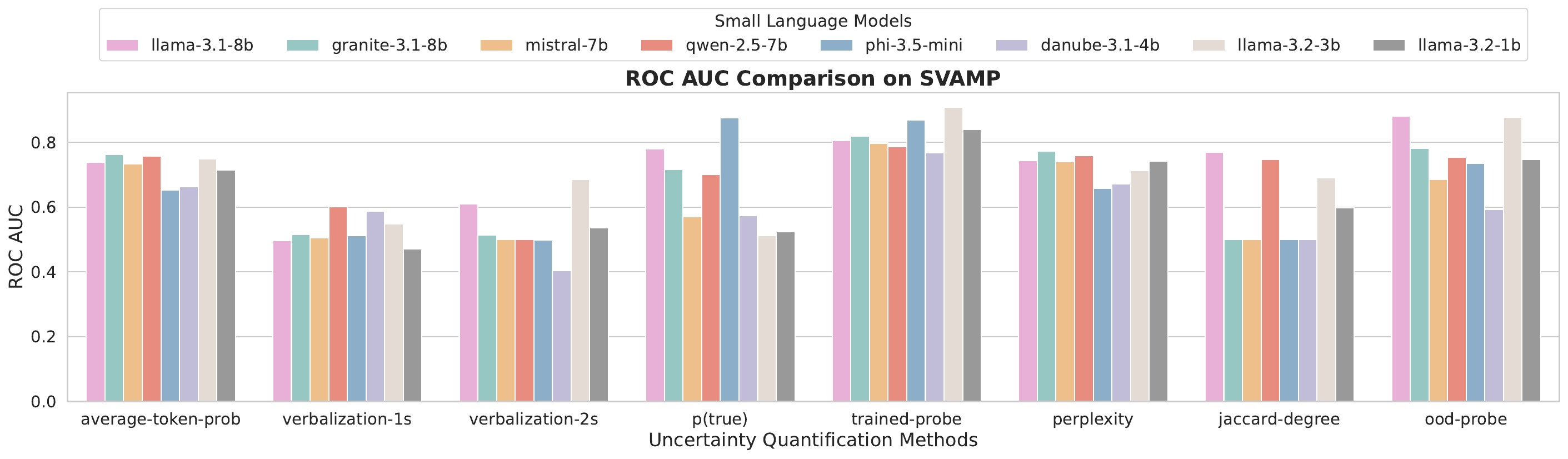}
    \vspace{-3mm}
    \caption{The ROC AUC scores measure the alignment between confidence and correctness across different SLMs and uncertainty quantification methods on CommonsenseQA, SVAMP, TruthfulQA, and WinoGrande. A higher ROC AUC indicates a stronger alignment.}
\label{fig:uq_align2} 
\end{figure*}

\clearpage
\subsubsection*{Results of Alignment between uncertainty and correctness.}
All the experiments shown on this page are conducted under CommonsenseQA, TruthfulQA, and WinoGrande datasets with all $8$ UQ methods.
\begin{figure*}[h!]
\centering
    \includegraphics[width=0.9\textwidth]{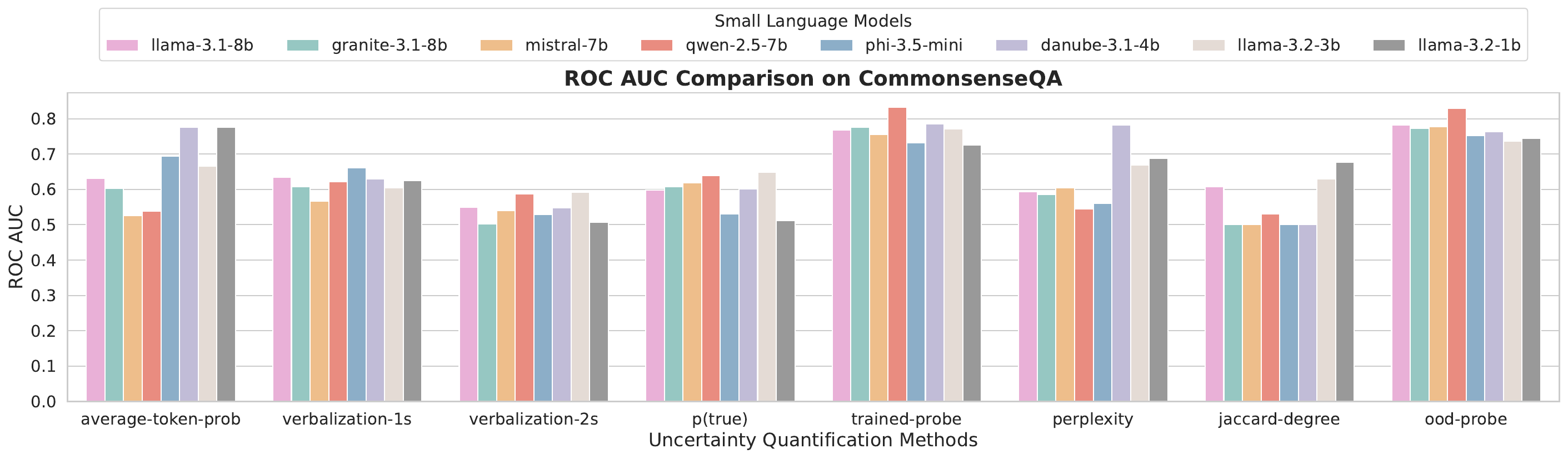}
    \includegraphics[width=0.9\textwidth]{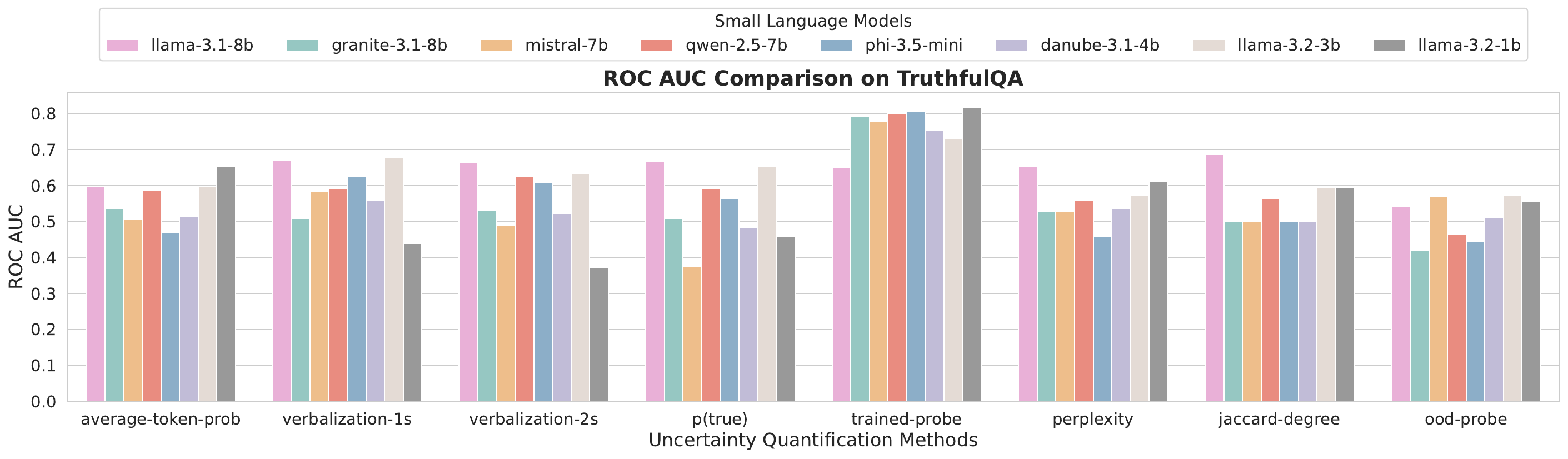}
    \includegraphics[width=0.9\textwidth]{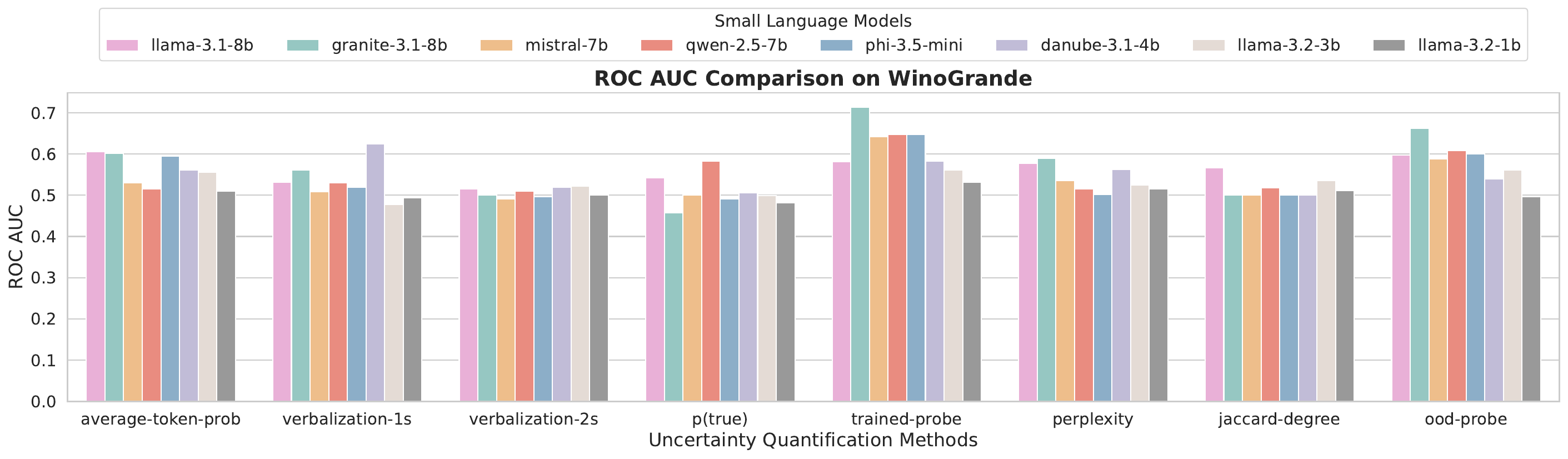}
    \vspace{-3mm}
    \caption{The ROC AUC scores measure the alignment between confidence and correctness across different SLMs and uncertainty quantification methods on CommonsenseQA. A higher ROC AUC indicates a stronger alignment.}
\label{fig:uq_align3} 
\end{figure*}

\clearpage
\subsection{Evaluation on Uncertainty-based Routing Approaches} \label{apdx:routing}

\subsubsection*{Results of routing to GPT-4o-Mini}
All the experiments shown on this page are conducted under GSM8K, MMLU, CommonsenseQA, TruthfulQA, CoQA, BoolQ, and OpenBookQA datasets with all $8$ SLMs.

\begin{figure*}[!h]
    \centering
    \includegraphics[width=1.\linewidth]{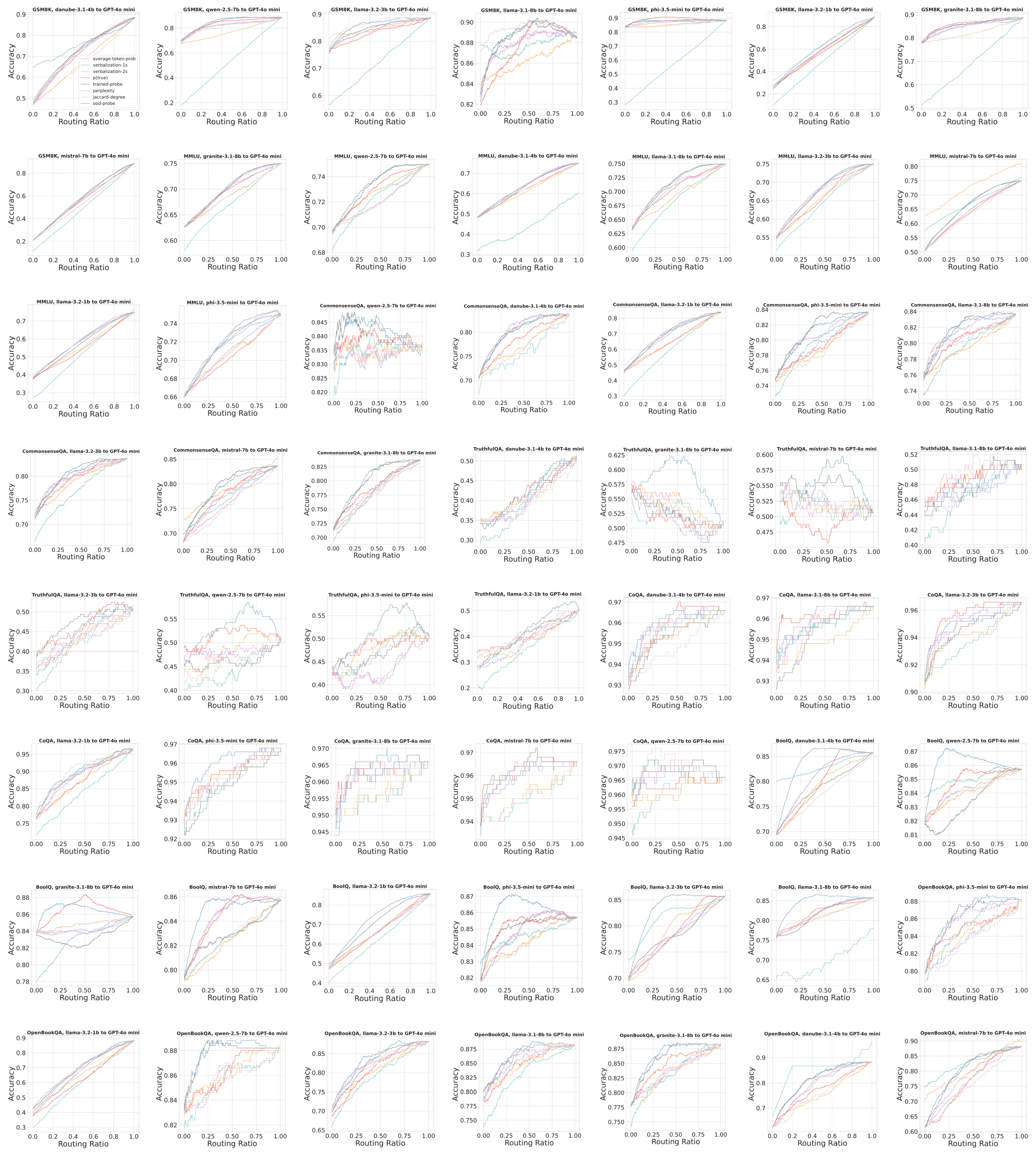}
    \caption{Overall accuracy vs. routing ratio (to GPT-4o-mini) with different UQ methods and SLMs.}
    \label{fig:route1}
\end{figure*}

\clearpage
\subsubsection*{Results of routing to GPT-4o-Mini}
All the experiments shown on this page are conducted under GSM8K, MMLU, CommonsenseQA, TruthfulQA, CoQA, BoolQ, and OpenBookQA datasets with all $8$ SLMs.
\begin{figure*}[!h]
    \centering
    \includegraphics[width=1.\linewidth]{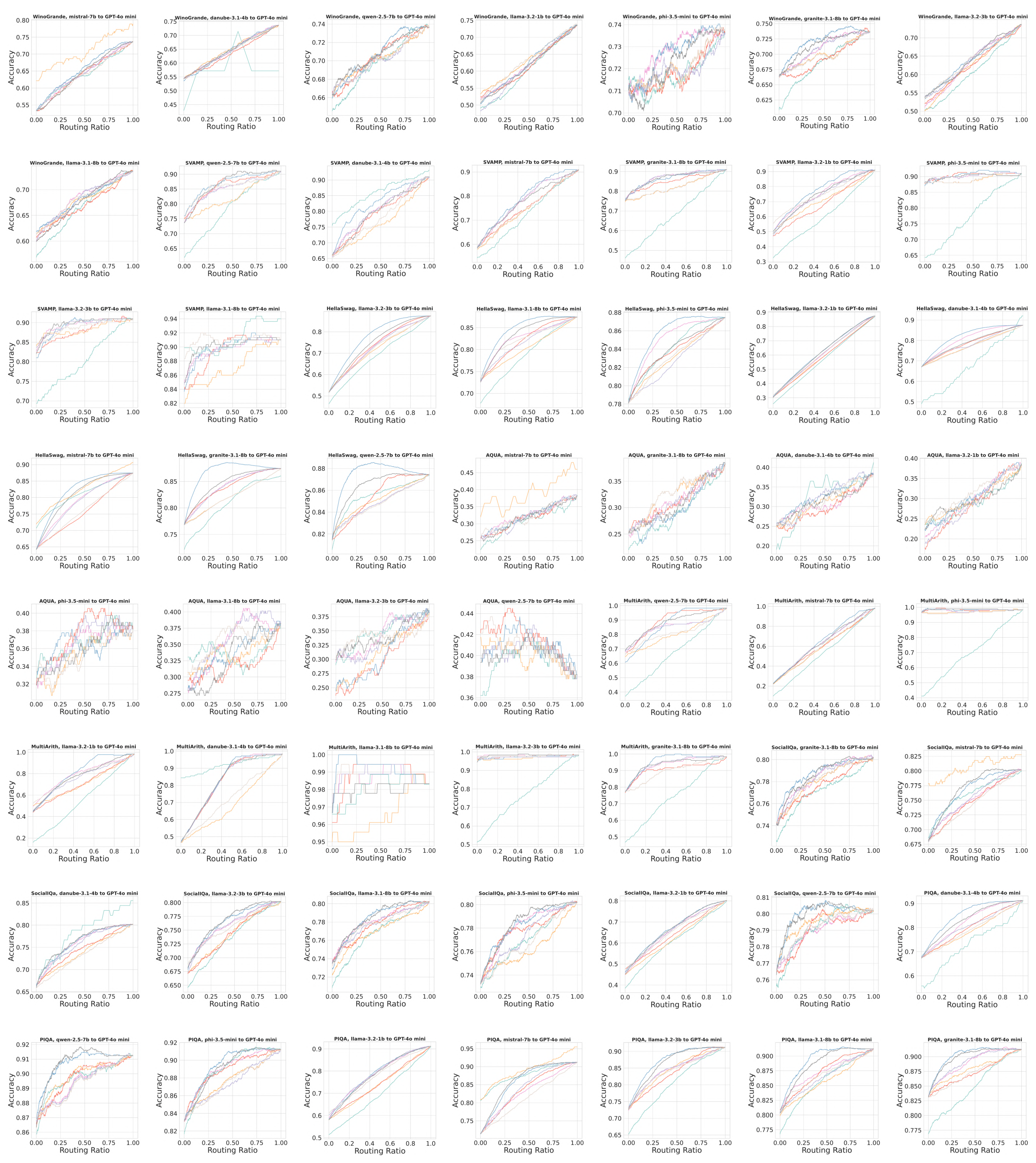}
    \caption{Overall accuracy vs. routing ratio (to GPT-4o-mini) with different UQ methods and SLMs.}
    \label{fig:route2}
\end{figure*}

\clearpage
\subsubsection*{Results of routing to Llama3.1-70B}
All the experiments shown on this page are conducted under GSM8K, MMLU, CommonsenseQA, TruthfulQA, CoQA, BoolQ, and OpenBookQA datasets with all $8$ SLMs.

\begin{figure*}[!h]
    \centering
    \includegraphics[width=1.\linewidth]{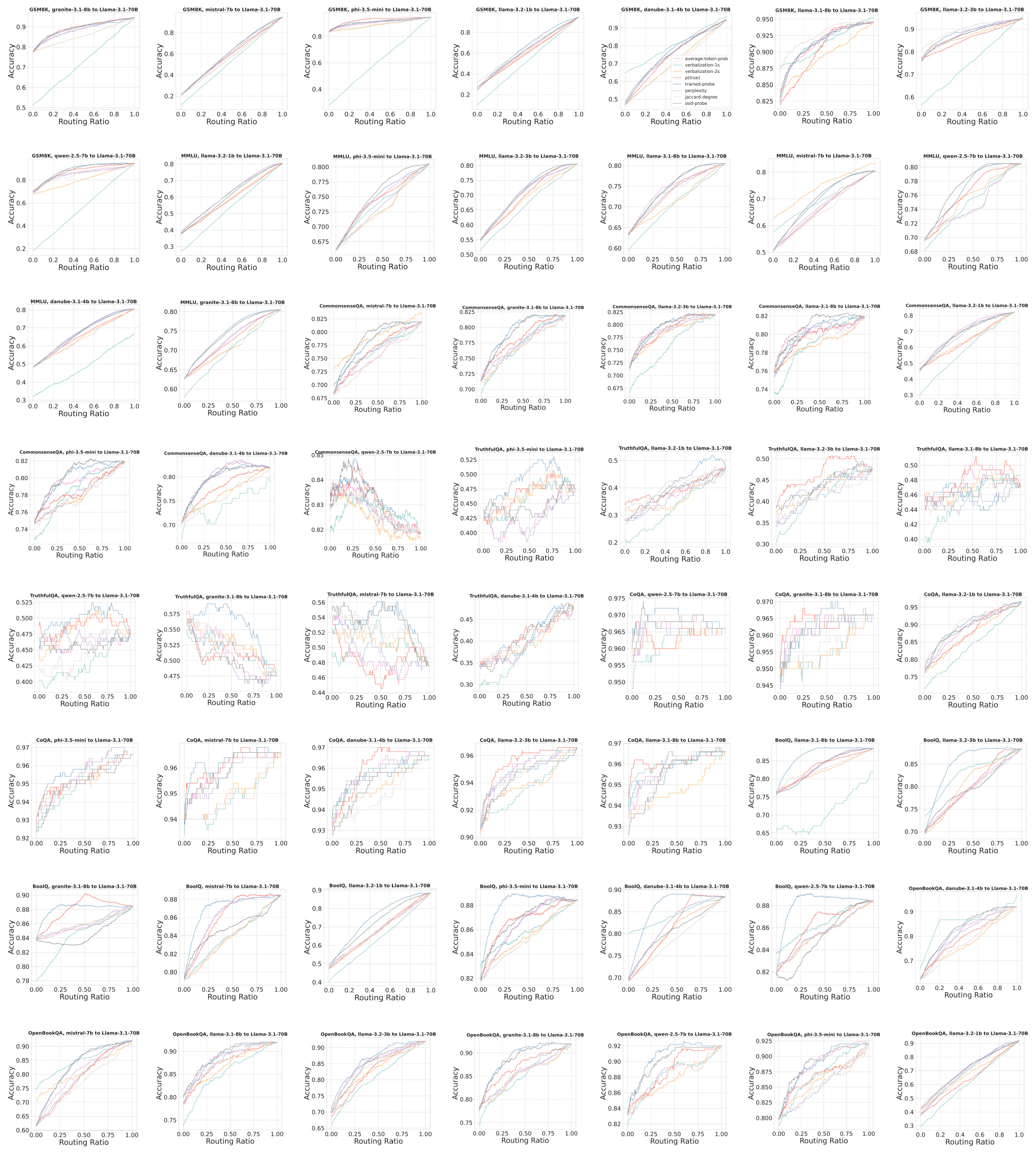}
    \caption{Overall accuracy vs. routing ratio (to Llama-3.1-70B) with different UQ methods and SLMs.}
    \label{fig:route3}
\end{figure*}

\clearpage
\subsubsection*{Results of routing to Llama3.1-70B}
All the experiments shown on this page are conducted under WinoGrande, SVAMP, HellaSwag, AQUA, MultiArith, SocialiQA, and PIQA datasets with all $8$ SLMs.
\begin{figure*}[!h]
    \centering
    \includegraphics[width=1.\linewidth]{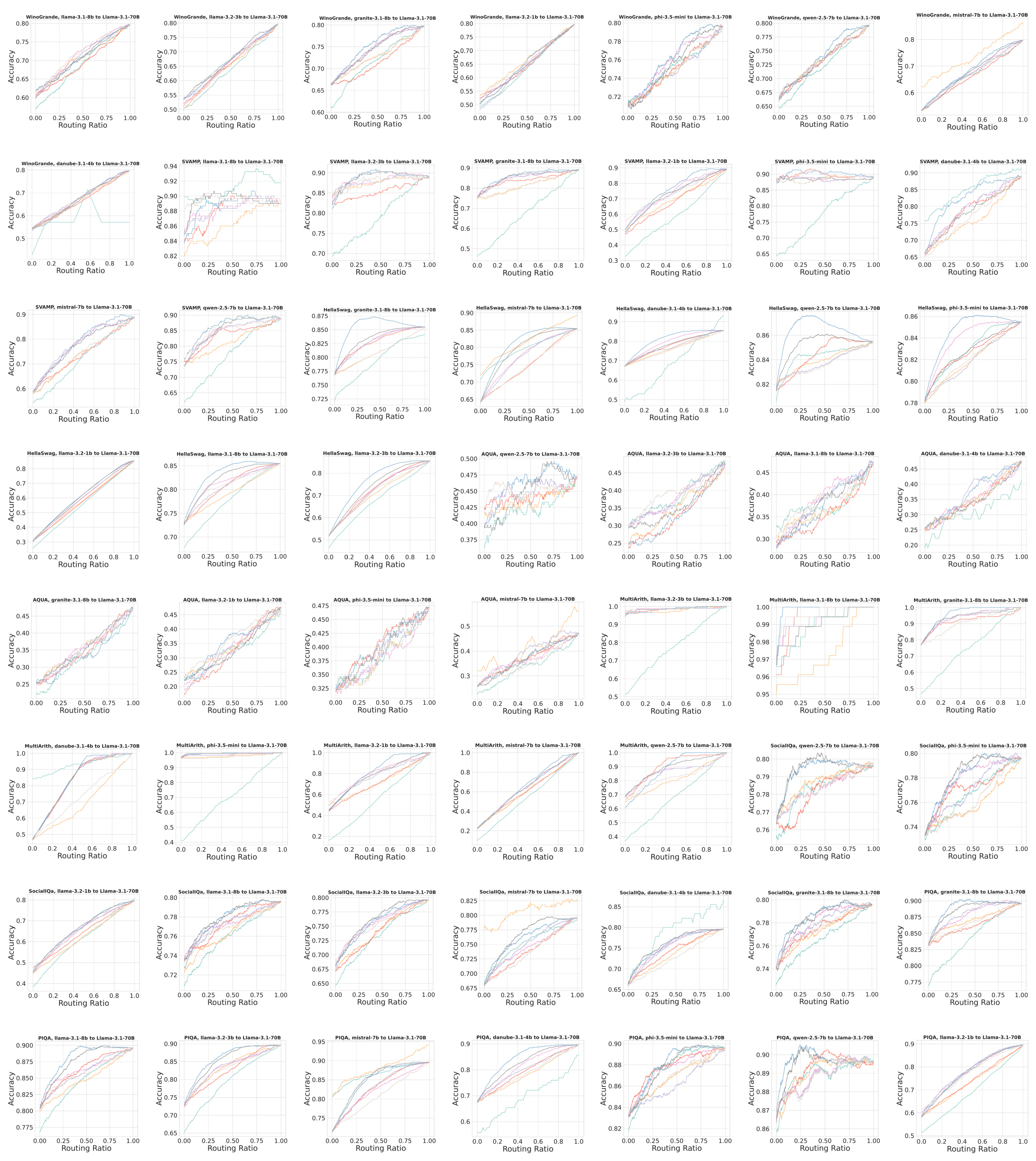}
    \caption{Overall accuracy vs. routing ratio (to Llama-3.1-70B) with different UQ methods and SLMs.}
    \label{fig:route4}
\end{figure*}

\clearpage
\subsection{Generalization Ability of Calibration Data on New Downstream Scenario} \label{apdx:generalization}

\subsubsection*{Generalization results on routing to GPT-4o-Mini}
All the experiments shown on this page are conducted under GSM8K, MMLU, CommonsenseQA, TruthfulQA, CoQA, BoolQ, and OpenBookQA datasets with all $8$ SLMs.
\begin{figure*}[!h]
    \centering
    \includegraphics[width=1.\linewidth]{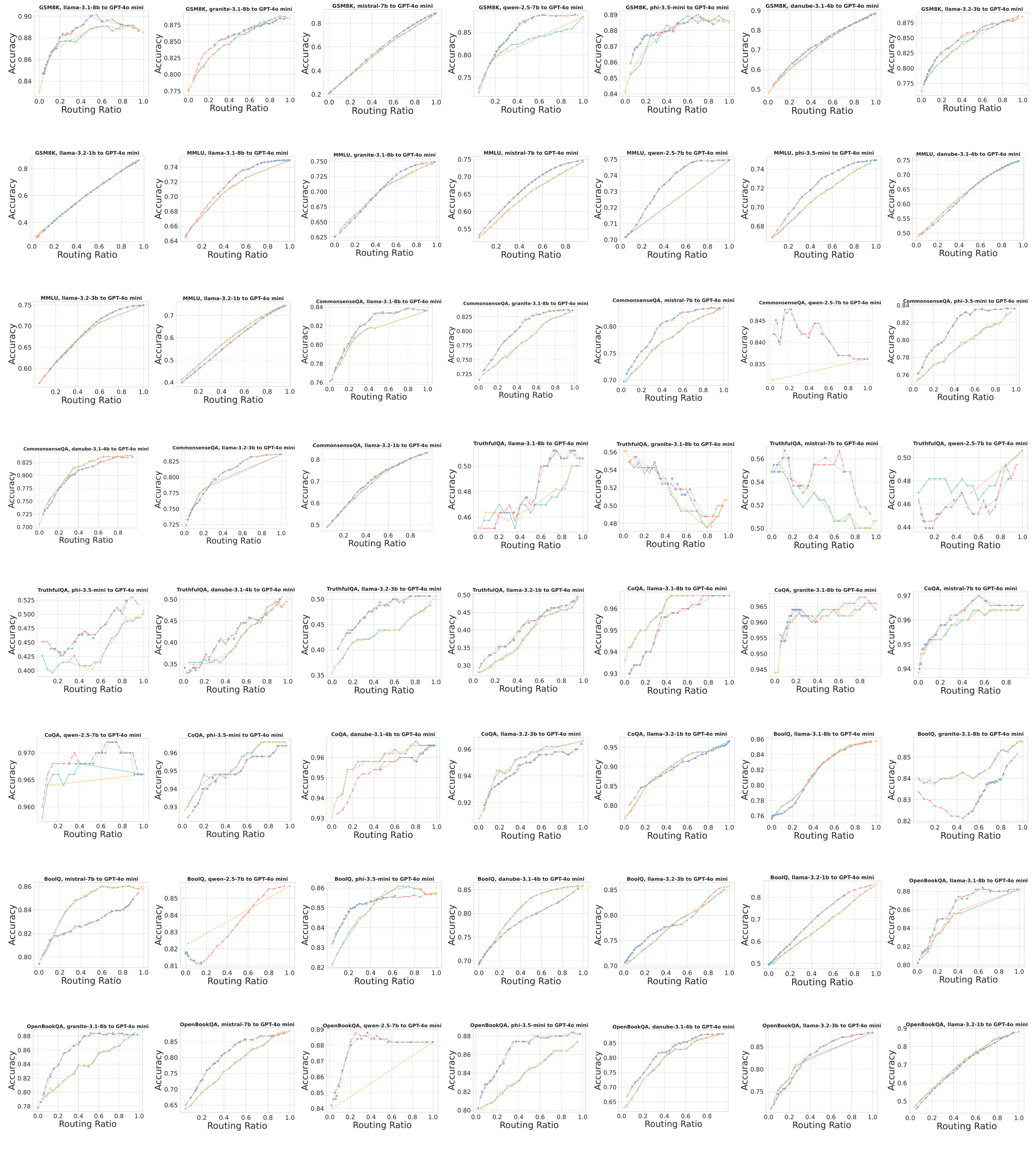}
    \caption{Assessing the generalization of calibration data to new downstream data for routing $8$ SLMs to GPT-4o-mini on $14$ datasets using two UQ methods (OOD Probe \& Perplexity). The legend in Figure~\ref{fig:cali_gen} is also used here.}
    \label{fig:add_cali}
\end{figure*}

\clearpage
\subsubsection*{Generalization Results on routing to GPT-4o-Mini}
All the experiments shown on this page are conducted under WinoGrande, SVAMP, HellaSwag, AQUA, MultiArith, SocialiQA, and PIQA datasets with all $8$ SLMs. 
\begin{figure*}[!h]
    \centering
    \includegraphics[width=1.\linewidth]{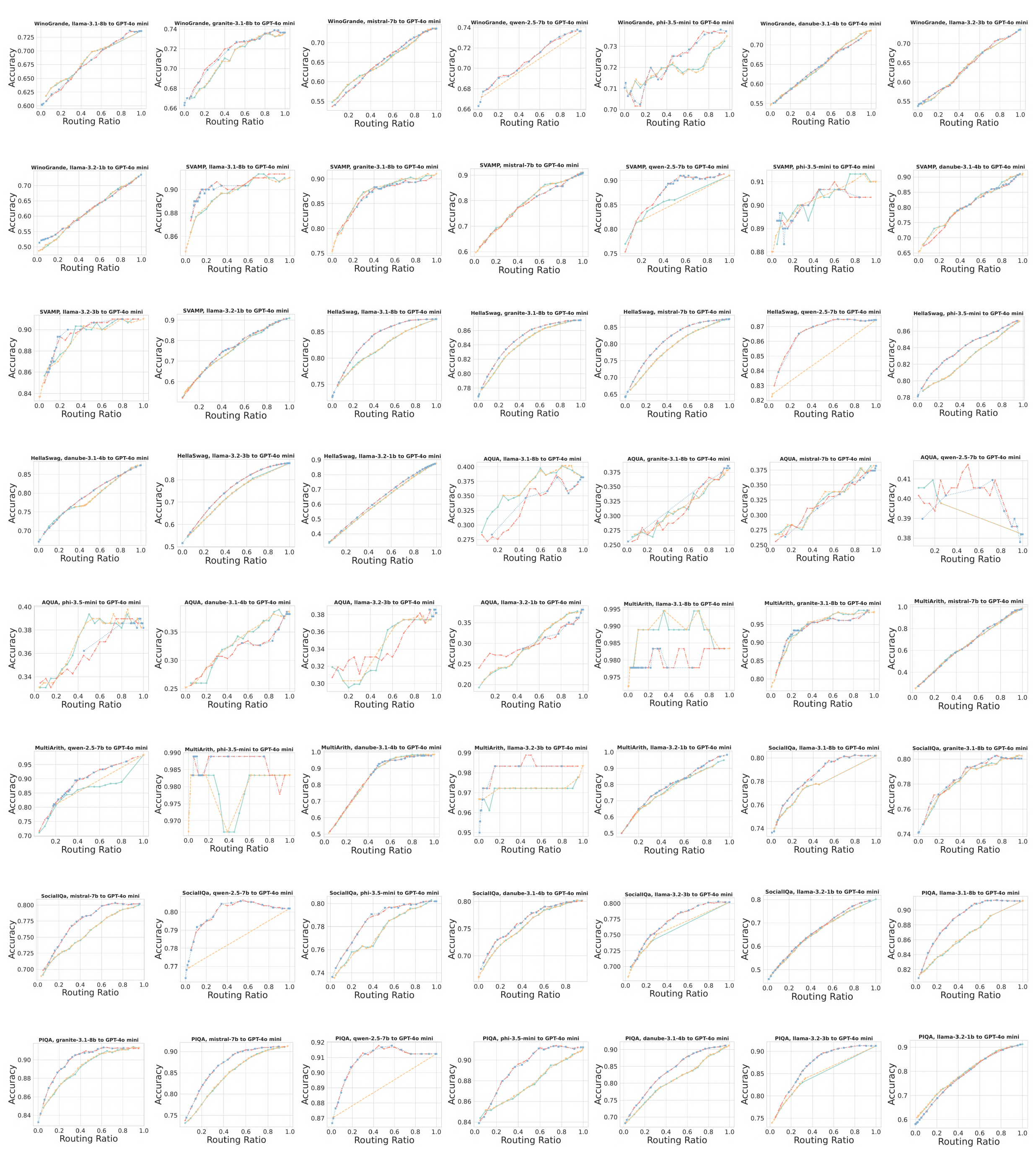}
    \caption{Assessing the generalization of calibration data to new downstream data for routing $8$ SLMs to GPT-4o-mini on $14$ datasets using two UQ methods (OOD Probe \& Perplexity). The legend in Figure~\ref{fig:cali_gen} is also used here.}
    \label{fig:add_cali1}
\end{figure*}

\clearpage
\subsubsection*{Generalization Results on routing to Llama3.1-70B.} 
All the experiments shown on this page are conducted under GSM8K, MMLU, CommonsenseQA, TruthfulQA, CoQA, BoolQ, and OpenBookQA datasets with all $8$ SLMs.
\begin{figure*}[!h]
    \centering
    \includegraphics[width=1.\linewidth]{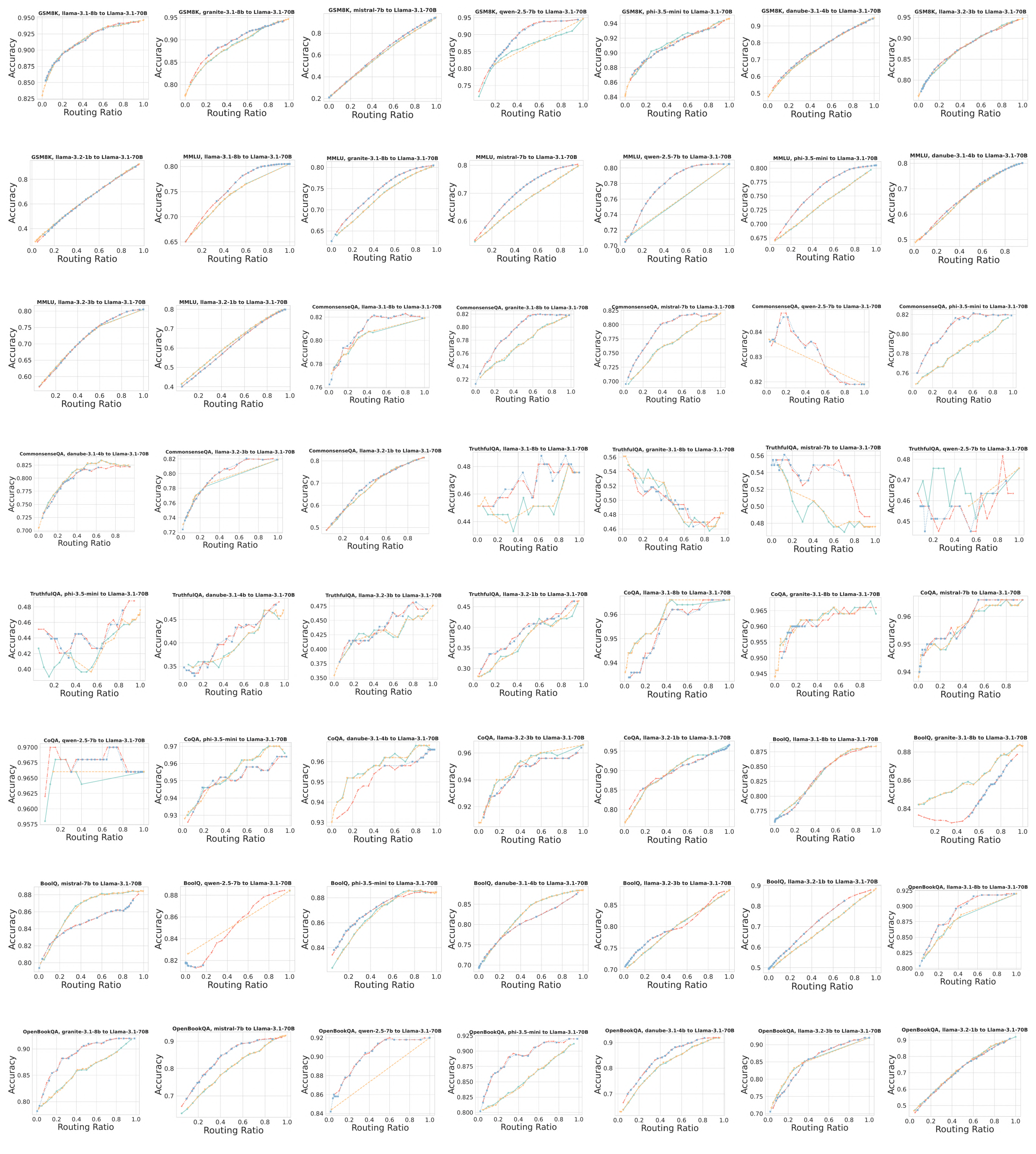}
    \caption{Assessing the generalization of calibration data to new downstream data for routing $8$ SLMs to Llama-3.1-70B on $14$ datasets using two UQ methods (OOD Probe \& Perplexity). The legend in Figure~\ref{fig:cali_gen} is also used here.}
    \label{fig:add_cali2}
\end{figure*}

\clearpage
\subsubsection*{Generalization Results on routing to Llama3.1-70B.} All the experiments shown on this page are conducted under WinoGrande, SVAMP, HellaSwag, AQUA, MultiArith, SocialiQA, and PIQA datasets with all $8$ SLMs. 
\begin{figure*}[!h]
    \centering
    \includegraphics[width=1.\linewidth]{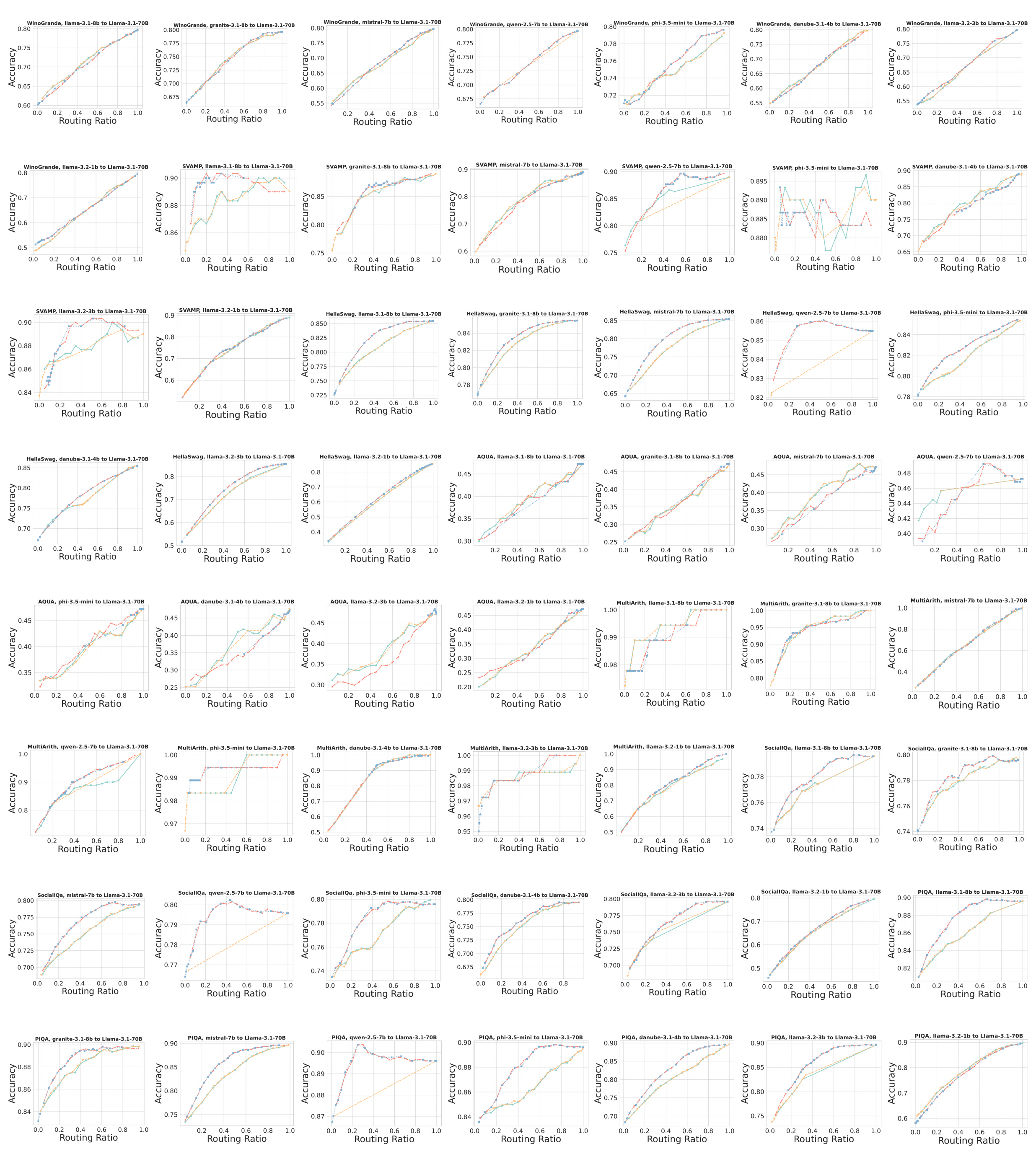}
    \caption{Assessing the generalization of calibration data to new downstream data for routing $8$ SLMs to Llama-3.1-70B on $14$ datasets using two UQ methods (OOD Probe \& Perplexity). The legend in Figure~\ref{fig:cali_gen} is also used here.}
    \label{fig:add_cali3}
\end{figure*}

\end{document}